\documentclass[10pt,journal,compsoc]{IEEEtran}

\usepackage{times}
\usepackage{soul}
\usepackage{url}
\usepackage{array}
\usepackage{graphicx}
\usepackage{subfigure}
\usepackage{amsmath}
\usepackage{amsfonts}
\usepackage[linesnumbered,ruled,vlined]{algorithm2e}
\usepackage{algorithmic}
\usepackage{multirow}
\usepackage{booktabs}
\usepackage{lineno}
\usepackage{pifont}
\usepackage{color}
\usepackage[justification=centering]{caption}
\usepackage{hyperref}
\usepackage{makecell}
\usepackage{mathtools}
\usepackage{colortbl}
\usepackage{makecell}
\usepackage{setspace}

\UseRawInputEncoding

\graphicspath{ {images/} }

\makeatletter
\renewcommand{\@thesubfigure}{\hskip\subfiglabelskip}
\makeatother
\ifCLASSOPTIONcompsoc
  \usepackage[nocompress]{cite}
\else
  \usepackage{cite}
\fi

\ifCLASSINFOpdf

\fi

\begin{document}
%
\title{GraphAttacker: A General Multi-Task Graph Attack Framework}
%
%

\author{Jinyin Chen, Dunjie Zhang, Zhaoyan Ming, Kejie Huang,~\IEEEmembership{Senior Member, IEEE}, { Wenrong~Jiang, and Chen Cui}
}

\markboth{IEEE TRANSACTIONS ON NETWORK SCIENCE AND ENGINEERING}%
{Shell \MakeLowercase{\textit{et al.}}: General Graph Attack Framework for Multi-Task}

\IEEEtitleabstractindextext{%
\begin{abstract}

Graph neural networks (GNNs) have been successfully exploited in graph analysis tasks in many real-world applications. {The competition between attack and defense methods also enhances the robustness of GNNs. In this competition, the development of adversarial training methods put forward higher requirement for the diversity of attack examples. {By contrast}, most attack methods with specific attack strategies are difficult to satisfy such a requirement. To address this problem, we propose GraphAttacker,} a novel generic graph attack framework that can flexibly adjust the structures and the attack strategies according to the graph analysis tasks. {GraphAttacker generates adversarial examples through alternate training on three key components: the multi-strategy attack generator (MAG), the similarity discriminator (SD),  and the attack discriminator (AD), based on the generative adversarial network (GAN).}
Furthermore, {we introduce a novel similarity modification rate (\emph{SMR}) to conduct a stealthier attack considering the change of node similarity distribution.  Experiments on various benchmark datasets demonstrate that GraphAttacker can achieve state-of-the-art attack performance on graph analysis tasks of node classification, graph classification, and link prediction, no matter the adversarial training is conducted or not.} {Moreover}, we also analyze the unique characteristics of each task and their specific response in the unified attack framework.{ The project code is available at https://github.com/honoluluuuu/GraphAttacker.}

%
\end{abstract}

}

\maketitle

\IEEEdisplaynontitleabstractindextext

\IEEEpeerreviewmaketitle

\IEEEraisesectionheading{\section{Introduction}\label{sec:introduction}}

\IEEEPARstart{O}{ur} lives are surrounded by various graphs,
such as social networks\cite{8411156}, e-commence networks\cite{wang2018billion}, biological networks\cite{8472785}, traffic networks\cite{latora2002boston}, {which produce abundant graph-structured data}. In recent years, a large number of GNNs\cite{hamilton2017inductive,kipf2017semi,Hu*2020Strategies} have been proposed to better analyze the various information contained in these data. Compared with the traditional graph embedding methods\cite{perozzi2014deepwalk,grover2016node2vec}, GNNs have achieved superior performance.

The development of GNNs has promoted wide applications of graph analysis in the real world tasks, i.e., node classification\cite{tang2015pte,wang2016linked}, graph classification\cite{ying2018hierarchical,lee2019self}, link prediction\cite{wang2017signed,tian2014learning}, and community detection\cite{7560670,7769223}. {The node classification task, learning hidden representation of nodes, is widely used in social user identification\cite{kipf2017semi} and financial scam detection\cite{pareja2020evolvegcn}. The graph classification task aggregates the node-level representation to the graph-level, {achieving} satisfying performance in biomolecular recognition\cite{ying2018hierarchical} and malware detection\cite{wang2019heterogeneous}. The link prediction task and the community detection task can predict the possible links in the future and divide similar nodes into the same community, respectively. They both play crucial roles in the personalized recommendation system\cite{8923033,8424608} of the e-commerce platform.}

\begin{table*}[]\setlength{\belowcaptionskip}{-0.7cm}\setlength{\abovecaptionskip}{0.2cm}
\center
\renewcommand\arraystretch{1.1}
\caption{{ Comparison of different attack methods. }}
\resizebox{150mm}{26mm}{

\begin{tabular}{ccc|ccc|cccc}
\hline
\multirow{3}{*}{{Method}} & \multirow{3}{*}{{Principle}} & \multirow{3}{*}{{Task} }                                        & \multicolumn{3}{c|}{{Attack strategy} }                                     & \multicolumn{4}{c}{Stealthiness constraints}                            \\ \cline{4-10}
                        &                            &                                                               & \makecell[c]{{Structure} \\{attack}} & \makecell[c]{{Attribute}\\ {attack}} & \multicolumn{1}{c|}{\makecell[c]{{Hybrid }\\{attack}}} & \makecell[c]{{Attack}\\ {budget}} & \makecell[c]{{Test} \\{statistic}}& \makecell[c]{{Similarity} \\{modification ratio}} & {L2-norm} \\ \hline
{DICE}                    & {Heuristic }                 & \makecell[c]{{Node classification,} \\{link prediction}}                          & {\checkmark}        &                 &                                   & {\checkmark}       &               &                              &         \\ \hline
{NETTACK }                & \makecell[c]{{Greedy} \\{algorithm} }          & {Node classification }                                          & {\checkmark}  & {\checkmark}      & {\checkmark}                                   & {\checkmark}           & {\checkmark}       &                             &         \\ \hline
{GF-Attack}               & \makecell[c]{{Approximate}\\ {spectrum }}      & {Node classification}                                           & {\checkmark}     &                 &                                   & {\checkmark}        &               &                              &         \\ \hline
{GUA }                    & \makecell[c]{{Classification}\\{ boundary}}    & {Node classification}                                           & {\checkmark }   &                 &                                   & {\checkmark }     &               &                             &         \\ \hline
{IG-JSMA  }               & {Gradients}                  & {Node classification}                                           &{ \checkmark }    & {\checkmark}    & {\checkmark}                      & {\checkmark}        &              &                             &         \\ \hline
{SGA }                    & {Gradients}                  & {Node classification}                                           & {\checkmark}      &                 &                                   & {\checkmark}       & {\checkmark}               &                             &         \\ \hline
{GraphAttacker }          & {GAN }                       & \makecell[c]{{Node classification,} \\{graph classification,} \\{link prediction}} & {\checkmark}   &{\checkmark}      & {\checkmark}                      & {\checkmark}      & {\checkmark}    & {\checkmark }                    & {\checkmark}          \\ \hline
\end{tabular}\label{com}}
\end{table*}

Since numerous GNNs have achieved satisfying performance in real-world tasks, the potential security issues of GNNs also pose severe threats to downstream applications. In social networks, malicious users may {threaten} social and political security by hiding their criminal behavior\cite{krebs2002mapping}. In various online data, de-anonymization attacks\cite{7938634,9101713} expose users {private information, leading to} privacy leakage. To reduce such threats as much as possible, both attack\cite{zugner2018adversarial,chen2018fast,chang2020restricted,xi2020graph,zhang2020backdoor,8714065,ma2020towards} and defense methods\cite{dai2019adversarial,8924766,zhang2019comparing,9305289,entezari2020all,tang2020transferring} are proposed to improve the robustness of GNNs. More specifically, existing attack methods can be mainly categorized as adversarial attacks\cite{zugner2018adversarial,chen2018fast,chang2020restricted} and poisoning attacks\cite{9006004,xi2020graph,zhang2020backdoor} to reveal the vulnerability of GNNs in the inference process and the training process, respectively. {Unfortunately, these attack methods can only attack a single graph analysis task and are difficult to apply directly to other tasks.} In the aspect of defense methods, they are brought up to improve the robustness of GNNs, such as adversarial training\cite{dai2019adversarial,8924766}, adversarial perturbation detectors\cite{zhang2019comparing,ioannidis2019graphsac}, graph purification defense\cite{entezari2020all}, and other defense methods\cite{9305289,tang2020transferring}.

In recent years, the defense methods on GNNs have effectively improved the robustness of graph analysis methods. Generating diversiform adversarial examples and imperceptible perturbations are still challenges for most attack methods. For the diversity of adversarial examples, most attack methods first obtain perturbation candidate sets through optimizers such as gradient learning\cite{chen2018fast}, evolutionary computation\cite{8714065}, meta-learning\cite{zugner2019adversarial}, and other optimizers\cite{chang2020restricted,zhang2020backdoor,8714065}. Then they select the perturbations with the most significant impact on the downstream graph analysis tasks from the candidate sets. These attacks are difficult to generate a diversiform perturbation candidate sets since the model parameters are fixed. {Although the existing attack methods have implemented graph structure attack\cite{chen2018fast,chang2020restricted}, feature attack\cite{9006004}, or hybrid attack\cite{zugner2018adversarial,zhang2020backdoor}, most of them aim at a specific graph analysis task, adopting the particular modification strategy and stealthiness constraint without considering the diversity of attack targets and strategy. For the perturbations' stealthiness, several works\cite{wu2019adversarial,zhang2020gnnguard,jin2020graph} point out that the key of an effective attack on the graph is to add links between dissimilar nodes. They have achieved satisfactory defense effects by deleting the links between dissimilar nodes.

In summary, it is still a challenge to bring up an attack method that is suitable for multiple graph analysis tasks and considers the diversity of adversarial examples. It is helpful to explore the generality and specificity of different graph analysis tasks, promoting defense research on the graph analysis tasks. For instance, in the blockchain transaction platform, phishing detectors classify the target address as normal or phishing. Recently, researchers modeled the blockchain phishing detector as different graph analysis tasks, including node classification\cite{wu2020phishers}, graph classification\cite{zhang2021blockchain}, and link prediction\cite{lin2020t} tasks. Designing a general attack framework can help evaluate the robustness of diverse phishing detectors under the same attack conditions and discover their common or unique vulnerabilities. This paper proposes GraphAttacker, a general multi-task graph attack framework, which can achieve diversiform adversarial attacks and state-of-the-art performance in various graph analysis tasks. The GAN-based adversarial example generator is suitable for various graph analysis tasks. It can overcome the limitation of perturbation generation from the candidate set to achieve diversiform attack targets and strategies.

Additionally, it is also convenient to carry out different stealthiness constraints after generating adversarial examples. However, is the existing stealthiness constraints imperceptible enough? We observed the deficiency of the current attack methods through the node similarity, which inspired us to design an \emph{SMR} to limit the adversarial perturbations. The main contributions of our work are summarized as follows:

\setlength{\hangindent}{2.3em}
$\vcenter{\hbox{\tiny$\bullet$}}$
We propose a general attack framework for multiple graph analysis tasks, namely GraphAttacker. It combines multiple modification strategies, attack scale, and stealthiness constraints according to different downstream tasks {and achieves efficient and general graph attacks in a self-adaptive strategy generation manner.}

\setlength{\hangindent}{2.2em}
$\vcenter{\hbox{\tiny$\bullet$}}$
Based on attack budget $\triangle$  and node degree distribution statistics $\Lambda$, our proposed \emph{SMR} ensures that the perturbations do not significantly change the similarity of the node pairs but further enhances the adversarial perturbations' stealthiness.

\setlength{\hangindent}{2.3em}
$\vcenter{\hbox{\tiny$\bullet$}}$
Extensive experiments conducted on node classification, graph classification, and link prediction demonstrate that GraphAttacker can achieve state-of-the-art attack performance under a certain task, no matter the adversarial training is conducted or not. Moreover, we also explore the unique characteristics of each task when conducting unified attacks and put forward some insights.

The rest of this paper is organized as follows. In Section~\ref{Sec2}, we introduce the related works on adversarial attacks and defenses on GNNs. In Section~\ref{Sec3}, we present the definition of several graph analysis tasks and our general attack problem. Section~\ref{sec4} describes our GraphAttacker for multi-task graph attack in detail. We illustrate and analyze the experimental results in Section~\ref{sec5}. In Section~\ref{sec6}, we conclude this paper and indicate the future work.

\section{RELATED WORK}\label{Sec2}
Our work builds upon three categories of recent research: graph attack, graph defense, and multi-task learning on graph.

\subsection{Graph Attack}
In this section, we briefly review graph attack methods, mainly categorized into node classification attacks, graph classification attacks, link prediction attacks, and community detection attacks.

\textbf{Node classification attack.} Zugner et al.\cite{zugner2018adversarial} proposed the first adversarial attack against the graph data to generate the adversarial examples iteratively, namely NETTACK. Zugner et al. further proposed Meta-Self\cite{zugner2019adversarial}, using meta-gradients to solve the bi-level problem underlying training-time attacks when the node classification model and its training weights are unknown. In \cite{wang2018attack}, Wang et al. used a greedy algorithm to generate malicious nodes' links and corresponding attributes. They designed a discriminator based on GAN to ensure that the perturbations are imperceptible. GF-Attack\cite{chang2020restricted} constructs the corresponding graph filter to realize the black box attack on the graph embedding method. GUA\cite{zang2020graph} searches for a group of anchor nodes that can affect the prediction results of the target node to achieve general attacks on node classification tasks. Both IG-JSMA\cite{wu2019adversarial} and SGA\cite{li2021adversarial} are gradient-based attack methods. They implement attacks based on the integral gradient information of the whole graph and the gradient information of the constructed subgraph, respectively.

\textbf{Graph classification attack.}} Several works have explored attacks on graph classification tasks. Dai et al. \cite{dai2018adversarial} regarded the decision-making process of the adversarial attack as a {Markov decision-making process (MDP). They proposed an attack method based on reinforcement learning (RL-S2V)} to study the attack on graph classification. Tang et al.\cite{tang2020adversarial} took the pooling operation of the {hierarchical graph convolution network (GCN) model as the attack target, thus preserving the wrong node information. PA-Backdoor\cite{zhang2020backdoor} and  GTA\cite{xi2020graph} are poisoning attacks on the graph classification task. They generate different subgraphs as backdoor triggers to achieve attacks in the training process.

\textbf{Link prediction and community detection attacks.} Chen et al. \cite{9141291} implemented {an iterative gradient attack (IGA)} by using the gradient information of the trained {graph auto-encoder (GAE)}\cite{kipf2016variational}. Chen et al. further considered the dynamic nature of real-world systems and proposed the first adversarial attack\cite{chen2019time} {for dynamic link prediction (DNLP)}. Besides the above graph analysis task, Chen et al. \cite{8714065} regarded community detection attacks as an optimization problem and proposed an attack strategy based on genetic algorithm and Q modularity. Yu et al.\cite{9172881} proposed an attack method {based on a genetic algorithm}, which disturbs the Euclidean distance between node pairs in the embedding space.

The current attack methods mainly adopt specific strategies to achieve attacks of a particular graph analysis task. These attack methods are difficult to apply to other graph analysis tasks directly. They cannot generate diversiform adversarial examples either. Moreover, the stealthiness of adversarial perturbations is also worthy of discussion.

\subsection{Graph Defense}

Since the attacks on the graph analysis tasks have been brought up, the defense research for the graph analysis tasks is also under intensive development.

\textbf{Adversarial training.}  As one of the most commonly used defense methods, adversarial training has achieved excellent defense performance against a series of similar adversarial examples. Dai et al.\cite{dai2019adversarial} designed perturbations with different L2-Norm constraints according to different positive target-context pairs and prevented perturbations from spreading on the graph through adversarial training. Feng et al.\cite{8924766} designed a virtual adversarial regularizer, which smooths the prediction distribution in the most sensitive direction to enhance the robustness of the model. Sun et al.\cite{sun2019virtual} added a virtual adversarial loss to the basic loss function of GCN to improve the generalization of GCN.

\textbf{Adversarial perturbation detectors.} The adversarial perturbation detectors can detect possible adversarial perturbations in advance by investigating the distinction between the adversarial examples and clean ones. According to the attack principle of NETTACK, Entezari et al.\cite{zhang2019comparing} captured the possible adversarial perturbations by calculating the mean of the KL divergences between the softmax probabilities of the node and its neighbor nodes. GraphSAC proposed by Ioannidis et al.\cite{ioannidis2019graphsac} randomly draws a subset of nodes and relies on graph-aware criteria to filter out the node-set contaminated by abnormal nodes. Wu et al.\cite{wu2019adversarial} measured the similarities between nodes by Jaccard index and removed the edges that connect very different nodes. GNNGuard\cite{zhang2020gnnguard} introduces the neighbor importance estimation and the layer-wise graph memory components, which learn how to assign higher weights to edges connecting similar nodes while pruning adversarial perturbations between unrelated nodes.

{\textbf{Other defense methods.} Besides adversarial training and adversarial perturbation detectors,} Chen et al.\cite{9305289} trained a distillation GCN model by using the output confidence of the initial GCN as a soft label. Based on graph purification defense, \cite{entezari2020all} performed {a low-rank approximation} on the graph to reduce the impact of NETTACK. Several defense methods\cite{tang2020transferring,zhu2019robust} distinguish between the adversarial examples and the clean ones through attention mechanism and train robust models by penalizing weights on adversarial nodes or edges. Bojchevski et al.\cite{bojchevski2019certifiable} increased the number of certifiably robust nodes through robust training, utilizing the worst-case margin to encourage the model to learn more robust weights.

\subsection{{Multi-task Learning on Graph}}
Multi-task Learning on Graph The existing multi-task learning in the graph field mainly focuses on learning node representations that can be used to perform multiple tasks\cite{DBLP:conf/aaai/LiuFDQC19,DBLP:conf/iclr/XuHLJ19,holtz2019multi}. Zhang et al.\cite{DBLP:conf/nips/ZhangWY18} exploited historical multitasking experience to learn how to choose a suitable multi-tasking model for a new task. MT-MVGCN\cite{huang2020multitask} extracts the abundant information of multiple views on a graph through a multi-view graph convolutional network but is limited to simultaneously tackling link prediction and node classification tasks. Buffelli et al.\cite{buffelli2020meta} proposed a more widely appliable meta-learning strategy for multi-task representation learning. They generated higher-quality node embeddings in the case of model-agnostic and task-agnostic.}

Combining multi-task learning and graph adversarial attacks can achieve a general attack on multiple graph analysis tasks with a single attack method. Analyzing the unique characteristics of each task and its specific response in the unified attack framework is beneficial to discover the vulnerabilities of different graph analysis tasks and promote defense research on them.

\section{PRELIMINARIES}\label{Sec3}
This section introduces the definition of several graph analysis tasks and the general attack problem. For convenience, the definitions of some necessary notations used in this paper are briefly summarized in TABLE~\ref{tab1}.

\begin{table}[tb]\setlength{\belowcaptionskip}{-0.5cm}\setlength{\abovecaptionskip}{0cm}
\centering
\caption{{ The definitions of notations. }}
\resizebox{8.5cm}{!}{
\begin{tabular}{lr}
 \arrayrulecolor{black}
 \toprule [1pt]

{Symbol}                  & {Definition} \\  \midrule
{$G=\{V,E,X\}$   }          &     { \makecell[r]{Input graph with nodes $V$, links $E$, and node attribute $X$} }     \\
{$G=(A,X)$ }              &      {\makecell[r]{Simplified graph of replacing the $V$ and $E$ \\ with the adjacency matrix $A$} }   \\
{$G_{\text {set}}=\left\{G_{1}, \cdots, G_{n}\right\}$ } & {Graph set consisting of $n$ graphs} \\
\specialrule{0em}{0pt}{1pt}
{$A^{\prime}_{c}(A^{\prime})$ } &   {Continuous(discrete) adjacency matrix with perturbations }   \\
\specialrule{0em}{1pt}{1pt}
{$X^{\prime}_{c}(X^{\prime})$ } &   {Continuous(discrete) node attribute with perturbations }   \\
\specialrule{0em}{1pt}{1pt}
{$G^{\prime}_{c}=(A^{\prime}_{c} ,X^{\prime}_{c})$} &    { Adversarial example with continuous values }       \\
\specialrule{0em}{1pt}{1pt}
{$G^{\prime}=(A^{\prime}£¬X^{\prime})$      }      &    { Adversarial example with discrete values  }     \\
\specialrule{0em}{1pt}{1pt}
{$\tau_{i} \in \mathcal{T}$           }    &     { Target instances, could be $G_i$,$v_i$,or $e_{ij}$ }  \\
{$G_{K-sub}=(A_{s u b}, X_{s u b})$ }  &     {\makecell[r]{$K$-hop neighbor subgraph constructed \\  according to the target instance $\tau_{i}$}} \\
{$\Psi(G^{'})$              }   &     { Set of generated adversarial examples}\\
{$N$                    } &    {Number of nodes on graph $G(G^{'})$    }  \\
{$f_{\theta}(\cdot)$     }                &   { Target GNN model with the parameters $\theta$    }   \\
{$F$                     }  &      {Category set of nodes on graph $G$   } \\
{$Y$                    }   &      {Category set of graphs on graph set $G_{set}$   }  \\
{$I$                     }  &      {Category set of links on graph $G$ }   \\
{$\mathcal{S}_{l}$ ($\mathcal{S}_{u})$   }    &    { Training (test) instances, could be $G_{set}$, $V$, or $E$}  \\
{$ y_{i} $            }   &     { Ground truth label of $\tau_{i}$   }\\
{$\mathcal{L}$         }    &     {Loss function of target GNN model $f_{\theta}(\cdot)$  }\\

{$Z$                    }   &     {Output of the target GNN model $f_{\theta}(\cdot)$   }  \\
{$H^{(l)}$              }   &     {The $l$-th hidden representation of the feature extractor  }       \\
{$W^{(l)}$              }   &    { The $l$-th trainable weight matrix of the feature extractor  }    \\
{$H$                    }   &    { Hidden layer dimension of feature extractor   }   \\
{$d$                   }    &    { Dimension of low-dimensional representation $z_i$   }  \\
{$W_{A}^{ex}$  and $W_{X}^{ex}$ }                  &     {  Dimension expansion matrices of $A$ and $X$ on MAG }   \\
{$k$ }                      &     { Attack scale on MAG      }   \\
{$W_{SD}^{(l)}$  and $b^{(l)}$}                   &      { The $l$-th trainable weight matrix and the bias of SD }   \\
\specialrule{0em}{1pt}{1pt}
{ $\mathcal{S}$, $f$, and $\sigma$ }                    &       {\makecell[r]{Sigmoid,  softmax, and relu active functions} }     \\
{$z_i$   }                  &     { Low-dimensional representation of node $v_i$ }  \\
{$s_{ij}$ }                 &    { Similarity score between nodes $v_i$ and $v_j$}\\
{$r$ }                      &     { Perturbation ratio on GraphAttacker  }    \\ \bottomrule

\end{tabular}}
\label{tab1}%
\end{table}

 A graph is represented as $G=\{V, E, X\}$ , where $V=\left\{v_{1}, \cdots, v_{n}\right\}$  is the node-set with $|V|=N$ , $e_{i, j}=<v_{i}, v_{j}>\in E$  denotes that there is a link between node $v_{i}$ and node $v_{j}$. The node topology of the graph is generally represented by the adjacency matrix $A \in\{0,1\}^{N \times N}$ , where  $A_{i, j}=1$ if node $v_{i}$  directly connected with $v_{j}$. $X \in\{0,1\}^{N \times D}$  is the node attribute matrix, {where $D$  denotes the dimension of $X$}. Generally, the adjacency matrix $A$  contains the information of $V$ and $E$ , so we use $G=(A, X)$  to represent a graph more concisely.

\setlength{\parskip}{0\baselineskip}
\noindent\textbf{DEFINITION 1 (Node classification).}
Given a graph $G=(A, X)$ and node-set $V=\left\{v_{1}, \cdots, v_{n}\right\}$, $V_{\text {l}}$, and  $V_{\text {u}} \subset V$ denote the labeled and unlabeled nodes, respectively. $F=\left[y_{1}, \cdots, y_{|F|}\right]$ is the category set of nodes. The purpose of the node classification task is to train a model $f_{\theta}^{\text {node}}(\cdot)$ with a parameter of $\theta$ through $G$ and  $V_{\text {l}}$, and use the trained $f_{\theta}^{\text {node}}(\cdot)$ to predict the categories of $V_{\text {u}}$.

\noindent\textbf{DEFINITION 2 (Graph classification).}
In the graph classification task, model $f_{\theta}^{\text {graph}}(\cdot)$ takes the labeled graphs $G_{\text {l}} \subset G_{\text {set}}=\left\{G_{1}, \cdots, G_{n}\right\}$ and the category set of graphs $Y=\left[y_{1}, \cdots, y_{|Y|}\right]$ as input. The graph classification task aims to predict the categories of unlabeled graph $G_{\text {u}}$ through the trained $f_{\theta}^{\text {graph}}(\cdot)$.

\noindent\textbf{DEFINITION 3 (Link prediction).}
Different from the node classification task, the link prediction task replaces  $V_{\text {l}}$ and $F$ with labeled links $E_{\text {l}}\subset E$ and link categories $I=\left\{0,1\right\}$, and trains model $f_{\theta}^{\text {link}}(\cdot)$ to predict the categories of unlabeled links $E_{\text {u}}$.

\noindent\textbf{PROBLEM 1 (General attack).} For a given target instance set $\mathcal{T} \subseteq \mathcal{S}_{u}$, where $\mathcal{S}_{u}$ could be $V_u,G_u$ or $E_u$ for different graph analysis tasks, and $\mathcal{S}_{l}$ is the training instance. The attacker aims to maximize the loss of the target instances $\tau_{i}$ on $f_{\theta}(\cdot)$, causing $\tau_{i}$ to get the wrong prediction result. Generally, we can define the general graph attack as:
 \begin{equation}\label{1}
\begin{array}{l}
\operatorname{maximize}_{G^{\prime} \in \Psi(G)} \sum_{\tau_{i} \in \mathcal{T}} \mathcal{L}\left(f_{\theta^{*}}\left(G^{\prime}, X, \tau_{i}\right), y_{i}\right) \\
\text { s.t. } \theta^{*}=\underset{\theta}{\arg \min } \sum_{\tau_{j} \in \mathcal{S}_{l}} \mathcal{L}\left(f_{\theta}\left(G, X, \tau_{j}\right), y_{j}\right)
\end{array}
 \end{equation}
{where $\Psi(G)$ is the set of generated adversarial examples.}

\setlength{\parskip}{0\baselineskip}
\section{METHOD}\label{sec4}
\begin{figure}\setlength{\belowcaptionskip}{-0.5cm}\setlength{\abovecaptionskip}{0.2cm}\vspace{-1em}
  \centering
  \includegraphics[width=1.0\linewidth]{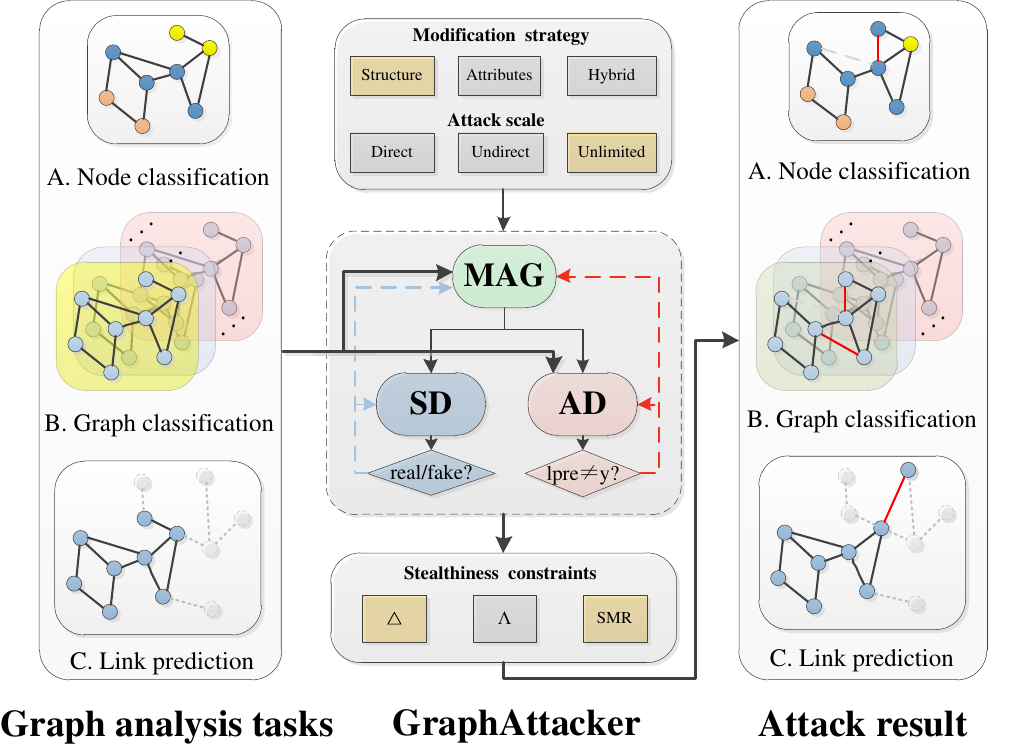}\\

  \caption{The framework of GraphAttacker. It generates adversarial examples through iterative training of three components, including multi-strategy attack generator (MAG), similarity discriminator (SD), and attack discriminator (AD). The example that satisfies both the attack effect and the stealthiness constraints are regarded as the final adversarial examples.}

  \label{fig:1}
\end{figure}

{Our proposed GraphAttacker conducts an attack on multiple graph analysis tasks by combining different modification strategies, attack scale, and stealthiness constraints. Fig.\ref{fig:1} shows the process of GraphAttacker on various graph analysis tasks. GraphAttacker framework consists of three modules: the multi-strategy attack generator (MAG), the similarity discriminator (SD), and the attack discriminator (AD).} GraphAttacker modifies the graph structure and node attributes within given stealthiness constraints to attack the corresponding target model through alternating training of MAG, SD, and AD. In this part, we take the node classification attack task as an example to introduce our GraphAttacker.

\subsection{Multi-Strategy Attack Generator (MAG)}
{Due to the discreteness of the graph-structured data, the existing graph attack methods prefer to selecting the optimal attack operation based on different optimizers, which are usually designed by the specific graph analysis task. The optimizers of these methods play a decisive role in generating adversarial examples and may have higher requirements on the model structure. When applied to other graph analysis tasks, these generation processes cannot be directly applied to other functions because they are not designed in a transferable manner.}

{Considering the diversity of attack targets, a generative adversarial example generator is worth considering. In our GraphAttacker, SD and AD can be regarded as our optimizers, which only play a guidance role in the generation process of MAG. Even if we can only obtain the prediction confidence of the target model without knowing the knowledge of the model structure, we can also guide the MAG to generate adversarial examples. This allows our MAG to generate adversarial examples for different graph analysis tasks quickly. Here we introduce the structure of MAG, multiple attack strategies, and multiple stealthiness constraints.}

\setlength{\parskip}{0\baselineskip}
\subsubsection{Structure of MAG}\label{sec4.4.1} {The proposed MAG generates diversiform adversarial examples through multiple attack strategies and stealthiness constraints.} MAG contains two modules, {including a feature extractor and a graph reconstructor.}

\noindent\textbf{Feature extractor.} To better learn the feature of graph structure and node attributes, we consider using the graph convolutional layer to extract the information {on the graph}. Here, the hidden layer $l + 1$ is defined as
\begin{equation}\label{2}
  H^{(l+1)}=\sigma(\hat{A} H^{(l)} W^{(l)})
\end{equation}

\noindent where $\hat{A}=\tilde{D}^{-\frac{1}{2}} \tilde{A} \tilde{D}^{-\frac{1}{2}}$, \emph{A} is the adjacency matrix, and $\tilde{A}=A+I_{N}$ is the adjacency matrix of the real graph \emph{G} with self-connections.  $I_{N}$ is the identity matrix and $\tilde{D}_{i i}=\sum_{j} \tilde{A}_{i j}$  denotes the degree matrices of $\tilde{A}$. In the first layer, we use the node attributes as input, i.e., $H^{(0)}= X$. $\sigma$ is the Relu active function. {Specifically, we consider a two-layer GCN\cite{kipf2017semi} as the graph feature extractor.} It maps the graph structure and node attribute information to a $d$-dimensional {node representation $Z$}, which is defined as:

\begin{equation}\label{3}
  Z=f^{node}(X, A)=f\left(\hat{A} \sigma\left(\hat{A} X W^{(0)}\right) W^{(1)}\right)
\end{equation}
{\noindent where $W^{(0)} \in \mathbb{R}^{N \times H}$ and $W^{(1)} \in \mathbb{R}^{H \times d}$ denote the trainable weight matrix of the hidden layer and the output layer with the hidden layer dimension $H$, respectively. The values of $H$ and $d$ determine the quality of the learned low-dimensional representation $Z$, and they are usually selected based on experimental results.} \emph{f} is the softmax function.

\begin{figure*}[htbp]\setlength{\belowcaptionskip}{-0.5cm}\setlength{\abovecaptionskip}{0.2cm}\vspace{-1em}
  \centering
  \includegraphics[width=.9\linewidth]{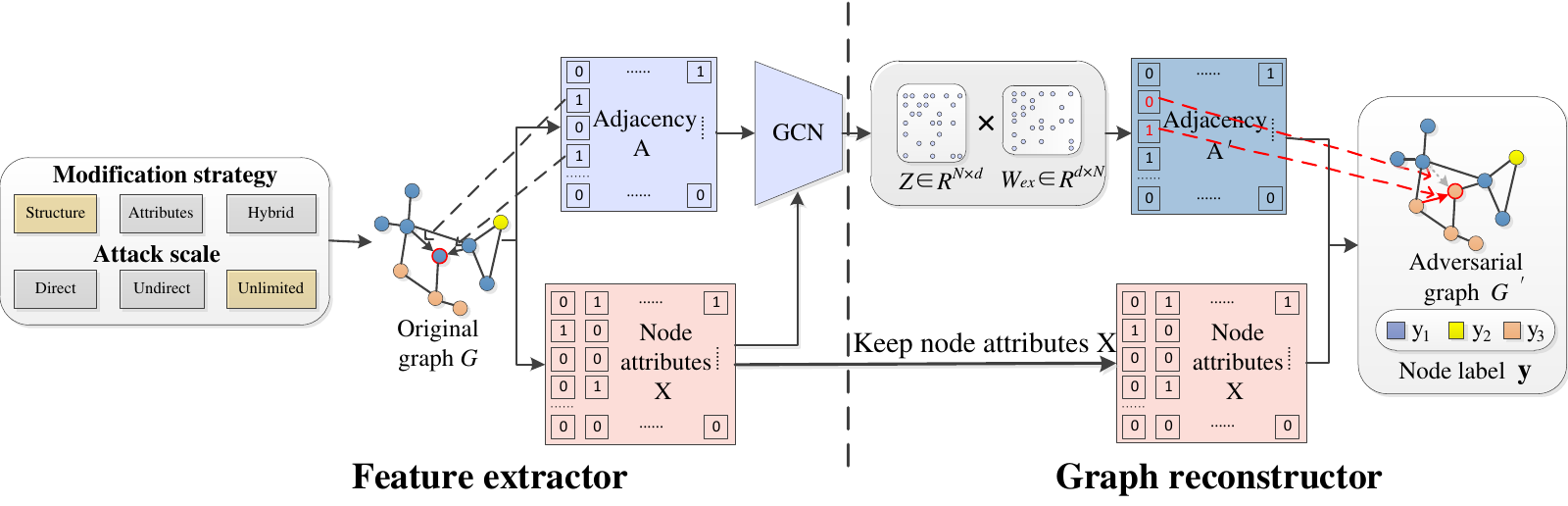}
     \caption{The multi-strategy attack generator (MAG) in node classification attack. The solid red line indicates the added links, and the dotted gray line indicates the deleted links. Here we choose an unlimited attack on the graph structure. We obtain the low-dimensional features of the graph through the feature extractor and then get our adversarial graph structure through the graph reconstructor.}\label{fig.2}
\end{figure*}

\noindent\textbf{Graph reconstructor.} {After obtaining the node representation \emph{Z} through the graph feature extractor}, we use a dimension expansion matrix $W_{e x}$ to reconstruct \emph{Z} into the continuous adversarial example $G_c^{'}$ :
\begin{equation}\label{4}
  G_{(c)}^{\prime}=\left\{\begin{array}{l}A_{(c)}^{\prime}=\mathcal{S}\left(\left(Z W_{e x}^{A}+\left(Z W_{e x}^{A}\right)^{T}\right) / 2\right) \\ X_{(c)}^{\prime}=\mathcal{S}\left(Z W_{e x}^{X}\right) \end{array}\right.
\end{equation}
where $Z \subset R^{N \times d}$. $W_{e x}^{A} \subset R^{d \times N}$ and $W_{e x}^{X} \subset R^{d \times D}$ are the dimension expansion matrices of graph structure \emph{A} and node attributes \emph{X}. Sigmoid function $\mathcal{S}$ maps the element values of generated data between [0,1]. Then we {obtain} discrete $G^{'}$ by the \emph{sign} function:

\begin{equation}\label{5}
  G^{\prime}=\left\{\begin{array}{l}A^{\prime}=\emph{sign}(A_{(c)}^{\prime}) \\ X^{\prime}=\emph{sign}(X_{(c)}^{\prime}) \end{array}\right.
\end{equation}
{where the \emph{sign} function sets the value greater than 0.5 to 1, and the others to 0.}

\subsubsection{Multiple attack strategies} Fig.\ref{fig.2} shows how MAG in GraphAttacker generates graph structure \emph{A} or node attributes \emph{X} based on different graph analysis attack strategies. {In our GraphAttacker, we design optional modify strategies and attack scale to meet the requirements of different attack scenarios.}

\noindent\textbf{Modify strategy.} {In MAG, the hybrid attack can usually achieve the most effective attack. In this scenario, MAG can adaptively select the optimal attack target (graph structure $A$ or node attribute $X$) and perturbation size according to the guidance of AD. Considering the attack scenarios that can modify the graph structure or the node attributes, MAG also provides graph structure attack and node attribute attack. The different modifying strategies are described as follows:}

\setlength{\parskip}{0\baselineskip}
\emph{Graph structure attack}: {We use the $Z$ and the $W_{ex}^A$ to generate the adversarial adjacency matrix $A^{'}$ by Eq.\ref{4} and Eq.\ref{5}}, and preserve the original node attributes $X^{'}=X$.

\emph{Node attribute attack}: {We want to preserve the original adjacency matrix $A^{'}=A$ in some cases.} {We only use the $Z$ and the $W_{ex}^X$} to generate the adversarial node attributes $X^{'}$ by Eq.\ref{4} and Eq.\ref{5}.

\emph{Hybrid attack}: Combining graph structure attack and node attribute attack, we both generate the adversarial adjacency matrix $A^{'}$ and node attributes $X^{'}$ to realize an adaptive modification strategy.

\noindent\textbf{Attack scale.} {In addition to the multiple modification strategies, the different attack scales can also help generate diversiform adversarial examples. Unlike the adversarial example generation mode that modifies the original graph once in each iteration, the MAG needs to generate a whole graph. The process will consume considerable computing resources, although it helps generate diversiform adversarial examples and allows it to extend to multi-strategy analysis tasks. Therefore, it is necessary to reduce the attack cost of GraphAttacker.}

\setlength{\parskip}{0\baselineskip}

{Here}, we use $K$-hop subgraph instead of the original graph in the attack process to achieve an efficient and low-cost attack. {The increase of the $K$ value MAG will conduct attacks on a larger scale and increase the complexity of the MAG.} Specifically, we select the target node and its $K$-hop neighbors from the original graph \emph{G} to form a $K$-hop subgraph {$G_{K-s u b}=(A_{s u b}, X_{s u b})$}. To prevent the node categories in the subgraph from becoming too concentrated,  {$20\%$ of subgraph nodes are randomly selected from other categories and added to the $G_{K-s u b}$, converted to the adversarial subgraph $G_{K-s u b}^{\prime}=(A_{s u b}^{\prime}, X_{s u b}^{\prime})$} by MAG to have the same size as $G_{K-s u b}$. When the attack on $G_{K-s u b}$ is successful, we will replace the subgraph $G_{K-s u b}$ with $G_{K-s u b}^{\prime}$ in the original graph \emph{G} to obtain the adversarial example $G^{\prime}=\left(A^{\prime}, X^{\prime}\right)$.

{After solving the high complexity problem of MAG, we can achieve low-cost and effective attacks in different attack scales. Like the above-mentioned hybrid attack, the unlimited attack is also an adaptive attack, which can automatically select the optimal attack target among the direct or indirect neighbors of the target instance. The different attack scale $k \in N_{+}$ are described as follows:}

\setlength{\parskip}{0\baselineskip}
\emph{Direct attack ($k=1$)}: Only delete the existing links of the target node, add a new one, or modify the attributes of the target node.

\emph{Indirect attack ($k\leq K,\quad k\neq 1$)}: Delete or add links in the 2-to-$K$ hop node pairs except the target node {in the subgraph}, or modify the attributes of these nodes.

\emph{Unlimited attack ($k\leq K$)}: Combining the above two attack scales, delete or add links between any pair of nodes {in the subgraph}, or modify the attributes of any node.

\setlength{\parskip}{0\baselineskip}
\subsubsection{Multiple stealthiness constraints}
While achieving efficient and effective attacks, it is also necessary to ensure imperceptible perturbations in our adversarial examples. One choice is to limit the perturbation in generating adversarial examples by setting a stealthiness constraint loss. Since MAG generates adversarial examples through a generative strategy, the additional stealthiness constraint loss can only serve as a guiding role for MAG, i.e., it cannot ensure that the generated adversarial examples satisfy our stealthiness requirements.

{Based on the above consideration, we perform stealthiness constraint evaluation after an adversarial example has been generated. The evaluation is independent of the attack process, which can help us achieve diversiform stealthiness constraints more easily. Here, we consider the following stealthiness constraints:}
\setlength{\parskip}{0\baselineskip}

{\noindent\textbf{Attack budget $\triangle$.} It accumulates the adjacency matrix and node attributes changes} and limits them within the budget $\triangle$ to constrain the overall perturbation size of the adversarial example.

\setlength{\parskip}{0\baselineskip}
{\noindent\textbf{Test statistic $\Lambda$.}} Zugner et al.\cite{zugner2018adversarial} pointed out that the graph structure's most prominent characteristic is its degree distribution. They provide a test statistic $\Lambda$ to limit the changes of node degree distribution.

{\noindent\textbf{Similarity modification ratio (\emph{SMR)}.}} Link prediction task usually use similarity scores to infer the possibility of link existence\cite{lu2011link}. The higher the similarity score between two nodes indicates the higher probability that they are linked. In other words, the links that exist between node pairs with low similarity scores may be abnormal. Based on this speculate, we constrain the perturbations at the node similarity level. Here we consider using cosine similarity to measure the similarity score between nodes $v_i$ and $v_j$:
\begin{equation}\label{6}
  s_{i j} =\frac{z_{i} \cdot z_{j}}{\left\|z_{i}\right\|\left\|z_{j}\right\|}
\end{equation}
where $v_i$  is our target node and $v_j \in S_i$ is the node that {has a link} with $v_i$. $z_i$ and $z_j$ are the i-th and j-th rows of $Z$, denoting the node representation of node $v_i$ and node $v_j$, respectively.

Using these similarity scores, we calculate the average similarity between the target node and its connected nodes to estimate the difference between $G$ and $G^{'}$. It is worth noting that since the graph structure in $G^{'}$ has changed, we need to use $z_i^{'}$  and $z_i^{'}$ to calculate $s_{i j}^{'}$ in $G^{'}$. The $SMR$ of node $v_i$ is:

\begin{equation}\label{7}
   S M R_{i}=\frac{(\sum_{j \in S_{i}} s_{i j}) /\left|S_{i}\right|-(\sum_{j \in S_{i}^{'}} s_{i j}^{'}) /\left|S_{i}\right|}{(\sum_{j \in S_{i}} s_{i j}) /\left|S_{i}^{'}\right|}
\end{equation}

\subsection{Similarity Discriminator (SD)}
{The generated adversarial examples should be highly similar to the original ones. For this consideration, we design an SD, which aims to learn the difference between $G_{K - s u b}$ and $G_{K - s u b}^{'}$, and distinguish the two as much as possible. SD also provides feedback to the MAG and guides it to generate an adversarial subgraph similar to the original one.
Since the responsibility of SD is to ensure that the adversarial example is similar to the original one, it is independent of downstream graph analysis tasks. So we use a classic multi-layer perception (MLP) with one hidden layer as our SD, whose output layer is set to a one-dimensional sigmoid function. Briefly, the details of the SD can be formulated as follow:}

{\begin{equation}\label{8}
  SD(G)=\mathcal{S}\left(W_{S D}^{(1)}(W_{S D}^{(0)}h^{(0)}+b^{(0)})+b^{(1)}\right)
\end{equation}
where $\{W_{S D}^{(0)} \in \mathbb{R}^{N^2 \times H_{SD}},W_{S D}^{(1)} \in \mathbb{R}^{H_{SD} \times 1} \}$ and $\{b^{(0)}\in \mathbb{R}^{H_{SD}},b^{(l)}\in \mathbb{R}^{1}\}$ are the parameters of the hidden layer and the output layer, respectively. $H_{SD}$ is the hidden layer's dimension of SD. The input $h^{(0)}$ is the row-wise long vector reshape from the graph structure $A$ or the node attributes $X$ according to the modification strategy. $\mathcal{S}$ represents the sigmoid function
of the output layer.}


\subsection{Attack Discriminator (AD)} Through alternate training of MAG and SD, we can get an adversarial example similar to the real one, but this does not satisfy our attack requirement. {We add the AD to the game between MAG and SD to form a three-player game, making the generated adversarial examples have effective attack ability.}

{ As our attack optimizer, the responsibility of AD is to guide MAG to generate adversarial examples, which can mislead the target model. Therefore, we design AD according to different target models. It is worth noting that when the target model has been trained, we only design the feedback to MAG according to the prediction confidence output of the target model, without the knowledge about the internal structure and the parameter information of the model. Considering that the untrained target model, we adopt the GCN model as our AD in the node classification task, which has the same model structure as the feature extractor in section \ref{sec4.4.1}, so we denote AD as:}

{\begin{equation}
\setlength{\abovedisplayskip}{1pt}
  AD(G)=f^{node}(X, A)=f\left(\hat{A} \sigma\left(\hat{A} X W^{(0)}\right) W^{(1)}\right)
\end{equation}
AD obtains the category prediction confidence of the nodes according to the input graph $G$, here $d={|F|}$.}

\begin{figure}\setlength{\belowcaptionskip}{-0.5cm}\setlength{\abovecaptionskip}{0.2cm}\vspace{-1em}
  \centering
  \includegraphics[width=0.7\linewidth]{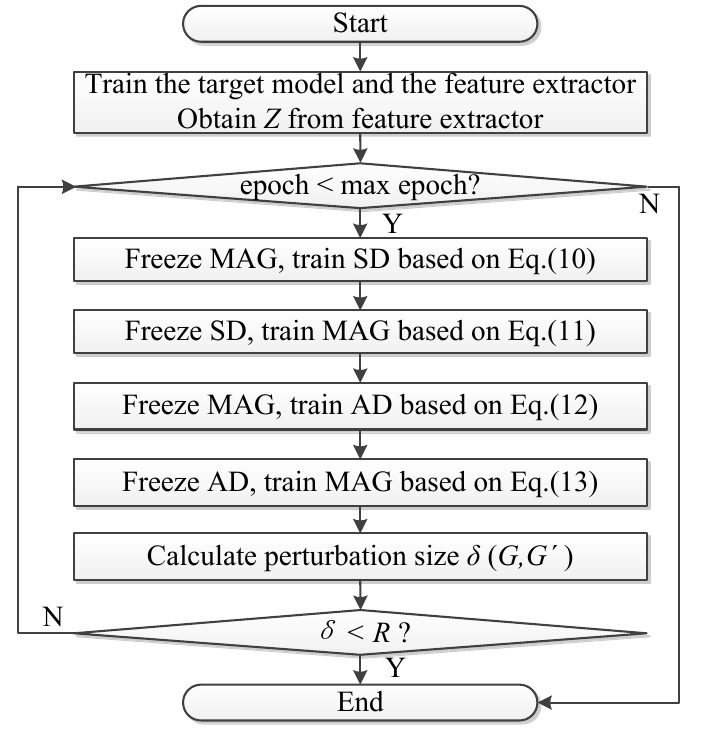}\\

  \caption{{The flow chart of three-player game training in GraphAttacker.}}

  \label{fig:lc}
\end{figure}

{\subsection{Training Procedure}\label{train}
GraphAttacker can continuously generate massive and diversiform adversarial examples during the testing process based on the proposed three modules of MAG, SD, and AD. During the generation process, we perform a three-player non-cooperative game among MAG, SD, and AD. As shown in Fig.\ref{fig:lc}, we design an alternate adversarial training process for MAG-SD and MAG-AD. SD and AD directly alternate training with MAG. SD and AD would indirectly affect each other through MAG. The three-player game of GraphAttacker includes the following four types:}

\setlength{\parskip}{0\baselineskip}

{\noindent\textbf{Train (MAG$\rightarrow$SD).}  Training SD via freezing the weights of MAG, and the optimization objective of SD is:
\begin{equation}\label                                                                                                                                                              {10}
  \begin{array}{c}
  \setlength{\abovedisplayskip}{1pt}
\setlength{\belowdisplayskip}{1pt}
\max \limits_{S D} \mathbb{E}_{G_{K\!-\!sub} \sim p_{\text {real}}}\left[\log S D\left(G_{K\!-\!s u b}\right)\right] \\
+\mathbb{E}_{\!G^{\prime}_{K\!-\!sub} \sim p_{M A G}\!}\left[\log \left(1-S D\left(G_{K\!-\!s u b}^{\prime}\right)\right)\right]
\end{array}
\end{equation}
where \!$G_{K\!-\!sub} \!\sim\! p_{real}$\! and \!$G_{K\!-\!sub}^{'} \!\sim\! p_{MAG}$\! denote original subgraph and adversarial subgraph generated by MAG, respectively.}

{\noindent\textbf{Train (SD$\rightarrow$MAG).}  Training MAG via freezing the weights of SD, and the optimization objective of MAG is:
\begin{equation}\label                                                                                                                                                              {11}
  \begin{array}{c}
\min \limits_{MAG} \mathbb{E}_{G^{\prime}_{K\!-\!sub} \sim p_{M A G}}\left[\log \left(1-S D\left(G_{K\!-\!s u b}^{\prime}\right)\right)\right]
\end{array}
\end{equation}}

{\noindent\textbf{Train (MAG$\rightarrow$AD).} Training AD via freezing the weights of MAG, and the optimization objective of AD is:
\begin{equation}\label                                                                                                                                                              {12}
  \begin{array}{c}
  \max \limits_{AD} \mathbb{E}_{G_{K\!-\!sub} \sim p_{real}}\left[\log \left(AD^{c}\left(G_{K\!-\!s u b}\right)\right)\right] \\
  + \mathbb{E}_{G^{\prime}_{K\!-\!sub} \sim p_{MAG}}\left[\log(1- \left(AD^{c}\left(G^{\prime}_{K\!-\!s u b})\right)\right)\right] \\
  s.t. \ c = argmax(y)
\end{array}
\end{equation}
where $y$ represents one-hot encoded ground-truth label for target instance, and $argmax(y)$ is the position of the maximum value in $y$. When the structure of the target model is not available, we can omit this training step and directly use the trained target model as our AD.}

{\noindent\textbf{Train (AD$\rightarrow$MAG).} Training MAG via freezing the weights of AD, and the optimization objective of MAG is:
\begin{equation}\label                                                                                                                                                              {13}
  \begin{array}{c}
  \min \limits_{MAG} \mathbb{E}_{G^{\prime}_{(c)K\!-\!sub} \sim p_{MAG}}\left[\log(1- \left(AD^{c^{'}}\left(G^{\prime}_{(c)K\!-\!s u b})\right)\right)\right] \\
  s.t. \ c^{'} = argmax(y_{tar})
\end{array}
\end{equation}
where $y_{tar}$ denotes the target attack label with one-hot encoded.}

The pseudo-code of the GraphAttacker training process is given in Algorithm \ref{algorithm1}. In each training epoch, it is necessary to ensure that the examples generated by the MAG are similar to the real ones before generating effective adversarial perturbations. As shown in Algorithm \ref{algorithm1}, GraphAttacker first trains SD for $k_{SD}$ times to improve the ability of SD to distinguish between the adversarial examples and the real ones (by Eq.\ref{10}). Then, we optimize $k_{MAG}$ times MAG through the trained SD and generate adversarial examples that are as similar as possible to the real ones (by Eq.\ref{11}). At last, we utilize the adversarial examples and the real ones to update the parameters of AD (by Eq.\ref{12}) and perform $k_{AD}$ times optimization on MAG under the guidance of AD to obtain adversarial examples with an effective attack ability (by Eq.\ref{13}).

\IncMargin{1em}
\begin{algorithm}
  \SetKwData{Left}{left}\SetKwData{This}{this}\SetKwData{Up}{up}
  \SetKwFunction{Union}{Union}\SetKwFunction{FindCompress}{FindCompress}
  \SetKwInOut{Input}{input}\SetKwInOut{Output}{output}

  \Input{Original graph $G=(A,X)$ or graph set $G_{set}=\left\{G_{1}, \cdots, G_{n}\right\}$, target instance $\tau$, stealthiness constraints $\mathrm{R}=\{\Delta, \Lambda, S M R\}$, perturbations $\delta(G, G^{'})$, the number of steps to apply to SD, MAG, AD, which are represented as $k_{SD}$, $k_{MAG}$ and $k_{AD}$}
  \Output{Adversarial examples, $G^{'}=(A^{'},X^{'})$}
  \BlankLine
  Train the target model:
  $\theta^{*} \leftarrow \min _{\theta} \mathcal{L}_{model}(\theta, A, X)$ and
  feature extractor on original graph $G$ to obtain $Z$ by Eq.\ref{3};
  Randomly initialize the dimension expansion matrix $W_{ex}$\;
  \For{number of training iterations}{
    \For{$k_{SD}$ steps}{\label{forins}
      {generate an adversarial example $G^{'}$ by Eq.\ref{4} and Eq.\ref{5}; update} the SD by ascending gradient:\\
      $\nabla_{\theta_{SD}} \left[\log \mathrm{SD}(G,\tau)+\log \left(1-\mathrm{SD}(G^{'},\tau)\right)\right]$\;
    }
    \For{$k_{MAG}$ steps}{\label{forins}
        update the MAG by descending gradient:\\
        $\nabla_{\theta_{MAG}} \log \left(1-\mathrm{SD}(G^{'},\tau)\right)$\;
    }
    \For{$k_{AD}$ steps}{\label{forins}
        update the AD by ascending gradient:\\
       {$\nabla_{\theta_{AD}} [\log \mathrm{AD}(G,\tau)+\log \left(1\!-\!\mathrm{AD}(G_{(c)}^{'},\tau)\right)]$};
        generate a new adversarial example $G^{'}$; {update} the MAG by descending gradient:\\
        $\nabla_{\theta_{MAG}} \log \left(1-\mathrm{AD}(G_{(c)}^{'},\tau)\right)$\;
    }
    generate a new adversarial example $G^{'}$\;
    \If{attack success}{\label{forins}
            \If{ $\delta(G, G^{'}) \textless \mathrm{R} $}{\label{forins}
                \textbf{break}\;

        }
    }
  }
  \textbf{return} The adversarial example $G^{'}$.
  \caption{\textbf{GraphAttacker: General multi-task graph attack}}\label{algorithm1}
\end{algorithm}

{\subsection{Multi-Task Graph Attack}} In GraphAttacker, the structure of MAG and SD is general to different graph analysis tasks. We can easily implement attacks on other graph analysis tasks by modifying AD {as a target graph analysis task model. Here, we briefly describe the difference in AD for graph classification and link prediction tasks.}

{\noindent\textbf{Graph classification attack.} In the graph classification task attack, the target instance is the whole graph $G$, and the expected output of AD is the graph-level label of the target instance. AD can be expressed as:
\begin{equation}
  AD(G)=f^{graph}(X, A)\in \mathbb{R}^{|Y|}
\end{equation}
where $Y=\left[y_{1}, \cdots, y_{|Y|}\right]$ denotes the category set of graphs, $|Y|$ is the number of graph categories.
After obtaining the category prediction confidence of graphs, GraphAttacker performs iterative training on MAG, SD, and AD as well as in Section\ref{train}.}

{\noindent\textbf{Link prediction attack.} The link prediction task takes the existence of the link instead of the node category as the prediction target. In this scenario, AD can be expressed as:
\begin{equation}
  AD(G)=f^{link}(X, A)\in \mathbb{R}^{N \times N}
\end{equation}
where $N$ is the total number of nodes in $G$. The $i$-th row and $j$-th column of AD indicate the category prediction confidence of the existence of the link between nodes $v_i$ and $v_j$. The training process of GraphAttacker is the same as above.}
\setlength{\parskip}{1\baselineskip}

\section{EXPERIMENTS} \label{sec5}
\setlength{\parskip}{0\baselineskip}
{This section applies GraphAttacker to different graph analysis tasks and compares} the achieved attack results with the baselines.
The main research questions of this section are:

\setlength{\hangindent}{3em}
\noindent\textbf{RQ1 } For a specific graph analysis task, how does the attack performance of GraphAttacker compared with existing attack methods?

\setlength{\hangindent}{3em}
\noindent\textbf{RQ2 } Will the perturbations generated by the existing attack methods significantly change the average similarity value of node pairs? If there is such a defect, can GraphAttacker make up for it?

\setlength{\hangindent}{3em}
\noindent\textbf{RQ3 } What is the impact of different graph datasets {on} attack performance?

\setlength{\hangindent}{3em}
\noindent\textbf{RQ4 }  What insights can we gain from the attack results on different graph analysis tasks through GraphAttacker?

\subsection{Experiment Setting} {We randomly split each dataset into the training set, validation set, and test set according to different division ratios introduced in different graph attack experiments. We select 20 instances from each class to form our whole target instance set $\mathcal{T}$.} For each attacked instance, we generate 20 adversarial examples. Once an adversarial example can successfully mislead the target graph analysis task, we consider the attack successful. At this time, we stop the attack process and output the adversarial example. {We conduct attack experiments on different ratios of training times for MAG, SD, and AD. The Adam optimizer is used to optimize our GraphAttacker, and the learning rate is searched in $[0.001,0.1]$. Other key hyper-parameters are set by a hyper-parameter search in the hidden layer dimension of feature extractor H$\in\{16,32,64,128,328\}$, the hidden layer dimension of SD $H_{SD} \in \{16,32,64,128,328\}$, the subgraph hop number $K\in \{1,2,3,4,5\}$, and the attack scale $k\in\{0,1,2,3,4,5\}$.}

{In our experiments, when the ratio of training times for MAG, SD, and AD is 1:1:1, the adversarial examples generated by GraphAttacker can both meet our attack requirements and stealthiness constraints. The final learning rate of MAG, SD, and AD is set to 0.03, and the hidden layer dimensions of the feature extractor and the SD are both set to 64. We implement our proposed GraphAttacker with Tensorflow}, and our experimental environment consists of i7-7700K 3.5GHzx8 (CPU), TITAN Xp 12GiB (GPU), 16GBx4 memory (DDR4) and Ubuntu 16.04 (OS).

\subsection{Datasets}
We evaluate GraphAttacker on 13 real-world datasets, including Cora\cite{mccallum2000automating}, Citeseer\cite{tarkowski2016closeness}, Pol.Blogs\cite{nagaraja2010impact}, Pubmed\cite{sen2008collective}, PROTEINS/PROTEINS-full\cite{borgwardt2005protein}, D\&D {(Dobson and Doig)}\cite{dobson2003distinguishing}, ENZYMES\cite{schomburg2004brenda}, NCI1 {(National Cancer Institute 1)}\cite{wale2008comparison}, IMDB-BINARY\cite{yanardag2015deep}, NS {(Network Science)}\cite{newman2006finding}, Yeast\cite{von2002comparative}, and Facebook\cite{mcauley2012learning}. We divide these datasets according to different graph analysis tasks.  {Among them, the citation datasets Cora, Citeseer, Pubmed, and the political blog dataset Pol.Blogs are usually applied to the node classification task. PROTEINS/PROTEINS-full, D\&D, ENZYMES, and NCI1 are bioinformatics datasets representing the relationship between biochemical structures, and IMDB-BINARY is a dataset containing movies classified by binary classification on IMDB. They are all used for the graph classification task. The NS, Yeast, and Facebook datasets used for the link prediction task contain the collaboration relationship between scientists in the network field, protein interaction relationship, and the user social relationship, respectively.} Some datasets do not have original node attributes, such as Pol.Blogs, NS, Yeast, and Facebook. In PROTEINS, DD, NCI1, and IMDB-BINARY, the node categories of each graph constitute their node attributes. The basic statistics of these datasets are summarized in TABLE\ref{tab:1}. {We show the average number of nodes and links in each dataset used in the graph classification task.}

\begin{table}[]\setlength{\belowcaptionskip}{-0.7cm}\setlength{\abovecaptionskip}{0.2cm}
\footnotesize
\centering
\caption{{Dataset statistics}}
\resizebox{88mm}{18mm}{
 \arrayrulecolor{black}
\begin{tabular}{c|cccccc}
\hline
Dataset       & Task  & \#Graph & .\# Nodes & .\# Links & .\# Classes & .\# Attributes \\ \hline
Pol.Blogs     & Node  & 1       & 1,490      & 19,090     & 2           & -              \\
Cora          & {Node/Link}  & 1       & 2,708      & 5,427      & 7           & 1,433           \\
Citeseer      & Node  & 1       & 3,312      & 4,732      & 6           & 3,703           \\
{Pubmed}        & {Node}  & {1}      &{19,717}     & {44,338}      &{3}            & {500} \\    \hline
PROTEINS-full & Graph & 1,113    & 39.06     & 72.82     & 2           & 29             \\
PROTEINS      & Graph & 1,113    & 39.06     & 72.82     & 2           & 1              \\
D\&D            & Graph & 1,178    & 284.32    & 62.14     & 2           & 1              \\
ENZYMES       &  Graph & 600     & 32.63     & 62.14     & 6           & 18             \\
{IMDB-BINARY}   & {Graph} & {1,000}    &{19.77}      &{96.53}     &{2}         & {1}   \\
NCI1          & Graph & 4,110    & 29.87     & 32.30      & 2           & 1              \\ \hline
NS            & Link  & 1       & 1,461      & 2,472      & 2           & -              \\
Yeast         & Link  & 1       & 2,375      & 11,693     & 2           & -              \\
Facebook      & Link  & 1       & 4,039      & 88,234     & 2           & -              \\ \hline
\end{tabular}
\label{tab:1}%
}
\end{table}

{\subsection{Baselines}
Since most of the existing graph classification attacks are poisoning attacks\cite{xi2020graph,zhang2020backdoor} or only attack the hierarchical graph pooling model\cite{tang2020adversarial}, so we only choose several attack methods in node classification and link prediction as our baseline. Among them, DICE, NETTACK and GF-Attack are baselines in node classification attacks. DICE can also be used for attack link prediction tasks, which form a link prediction attack baseline with IGA. For all the baselines, we use the attack results reported by the original authors when possible. When the baselines lack the attack results we need, we conduct experiments based on the source code released by the authors and their suggested parameter settings. The baselines are briefly described as follows:}

\setlength{\parskip}{0\baselineskip}
{\textbf{DICE}\cite{waniek2018hiding}: In node classification attack, DICE randomly disconnect $b$ links of the target node and then randomly connect the target node to $M-b$ nodes of different categories, where $M$ is the original number of links of the target node. In a link prediction attack, the nodes $v_i$ and $v_j$ corresponding to the target link $e_{ij}$ are used as target nodes.}

{\textbf{NETTACK}\cite{zugner2018adversarial}: NETTACK generates adversarial perturbations for graph structure and node attributes. {It ensures the perturbations are imperceptible by preserving the degree distribution and attributes co-occurrence probability.}

{\textbf{GF-Attack}: \cite{chang2020restricted}: GF-Attack attacks graph embedding models by constructing corresponding graph filters in a black box background.}

{\textbf{GUA}: \cite{zang2020graph}: GUA searches for a group of anchor nodes that can influence the prediction result of the target node and implements general attacks on the node classification task.}

{\textbf{IG-JSMA}: \cite{wu2019adversarial}: IG-JSMA introduces the integral gradient to solve the issue of inaccurate gradients on the discrete data.}

{\textbf{SGA}: \cite{li2021adversarial}: SGA focuses on reducing the complexity of calculating gradients on large-scale graph data. It attacks the subgraph with a simplified gradient approach.}

{\textbf{IGA}\cite{9141291}: IGA extracts the gradient matrix of the GAE model, which flips the link with the largest gradient in each iteration, and obtains adversarial examples through multiple iterations.}

\begin{figure*}[htbp]\setlength{\belowcaptionskip}{-0.5cm}\setlength{\abovecaptionskip}{0.2cm}\vspace{-1em}
  \centering
  \includegraphics[width=0.95\linewidth]{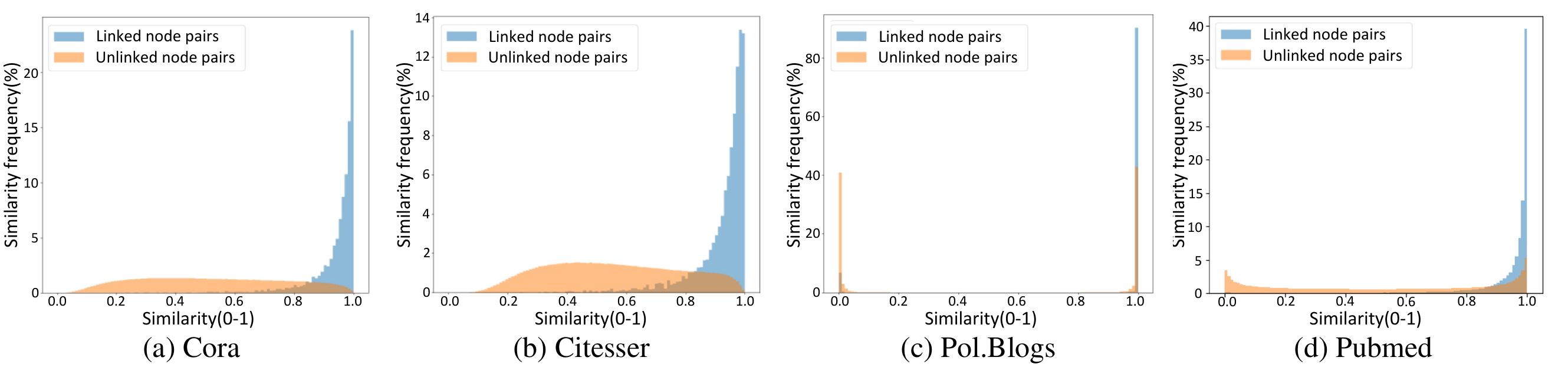}\\
  \caption{Node similarity distribution of different datasets reflects the similarity frequency between linked and unlinked node pairs. We can observe that more than $90\%$ of the linked node pairs have a similarity higher than 0.8.}
  \label{fig:3}
\end{figure*}

\subsection{Evaluation Metrics} To evaluate the performance of {the attacks}, we use the following four metrics:

\setlength{\hangindent}{2em}
$\vcenter{\hbox{\tiny$\bullet$}}$ \textbf{Attack Success Rate (ASR):} It represents the ratio of the targets which will be successfully attacked under given constraints, which is defined as:
\begin{equation}\label{17}
  \mathrm{ASR}=\frac{\text { Number of successful attacked instances }}{\text { Number of attacked instances }}
\end{equation}

\setlength{\hangindent}{2em}
$\vcenter{\hbox{\tiny$\bullet$}}$ \textbf{Average Modified Links (AML):} AML is designed for graph structure attack, which indicates the average number of modified links of each attack process.
\begin{equation}\label{18}
  \mathrm{AML}=\frac{\text { Number of modified links }}{\text { Number of attacked nodes }}
\end{equation}

\setlength{\hangindent}{2em}
$\vcenter{\hbox{\tiny$\bullet$}}$ \textbf{Average Modified Attributes (AMA):} Different from AML, AMA is designed for node attribute attacks, which indicates {each attack process's average attribute perturbation size.}
\begin{equation}\label{19}
  \mathrm{AMA}=\frac{\text { Number of modified attributes }}{\text { Number of attacked instances }}
\end{equation}

\setlength{\hangindent}{2em}
$\vcenter{\hbox{\tiny$\bullet$}}$ \textbf{L2-Norm $\|\cdot\|_{2}$:} In graph classification attack, the node attributes of PROTEINS-full and ENZYMES are continuous values. We use the L2-Norm to measure the magnitude of the perturbations in this case.
\begin{equation}\label{20}
\left\|X, X^{\prime}\right\|_{2}=\sqrt{\sum_{i}^{N} \sum_{j}^{D}\left(x_{i j}-x_{i j}^{\prime}\right)^{2}}
\end{equation}
where $X/X^{\prime}$ denotes the original/adversarial node attributes, $N$ is the number of nodes in the graph, $D$ is the dimension of the node attributes, and $x_{i j}/x_{i j}^{\prime}$ is the $j-$th original/adversarial node attribute value of the $i-$th node.

In our evaluation system, ASR reflects the effectiveness of different attack methods most intuitively. A higher ASR is equivalent to a stronger attack capability. However, attack performance does not only depend on the ASR, the stealthiness of perturbations is also crucial. AML is an evaluation metric used to evaluate the stealthiness of structural perturbations, while AMA and L2-norm are suitable for measuring perturbations in node attributes. In general, we expect to achieve higher ASR with the lowest possible AML, AMA, or L2-norm. Moreover, calculating the changes in node degree distribution and node similarity by disturbance is also an important approach to evaluate attack performance, which will be introduced in detail in sections \ref{5.5} and \ref{5.6}.

\subsection{Node Similarity Analysis}\label{5.5}
To explore the relationship between the similarity of node pairs and the probability of the existence of links, {we take the datasets of Cora, Citseer, Pol.Blogs, and Pubmed as examples to plot several figures about the similarity frequency between their linked and unlinked node pairs in Fig.\ref{fig:3}.} We use the low-dimensional representation of the nodes obtained from Eq.\ref{3} to calculate the cosine similarity (by Eq.\ref{6}) between node pairs. The x-axis is the similarity value of the node pairs, and the y-axis is the similarity frequency. It is evident that linked node pairs tend to have higher similarity, and the similarity of more than 90\% of linked node pairs is greater than 0.9. In other words, two nodes with low similarity are unlikely to be linked.

\setlength{\parskip}{0\baselineskip}
Based on this phenomenon, we conclude that we should pay more attention to {generating} effective adversarial perturbations between those nodes with high similarity. {Therefore, we restrict the \emph{SMR} to preserve the node similarity distribution of the original graph.} In the attack of node classification and link prediction, we construct the $K$-hop subgraph as the attack target to reduce the cost of the attack. We added two types of nodes in the process of constructing the $K$-hop subgraph. 1) \emph{Random nodes}: We randomly add some nodes with different categories to the original subgraph; 2) \emph{High similarity nodes}: Only the nodes {which have more than 0.9 similarities} with the target node are added to the original subgraph. In this way, we limit the influence of the perturbation candidate set on node similarity. Specifically, the number of added nodes is $20\%$ of the size of the original subgraph.

\subsection{ Multi-Task Graph Attack}\label{5.6}
 In this subsection, we mainly discuss question \textbf{RQ1}, and take node classification attack as an example to make a detailed analysis of question \textbf{RQ2}. To better demonstrate the graph attack performance of GraphAttacker in multi-task, we conducted attack experiments on tasks of node classification, graph classification, and link prediction.

\subsubsection{Node classification attack}\label{section:5.5.1}
Referring to some previous works\cite{zugner2018adversarial,zugner2019adversarial}, we first divide each dataset into three parts: 20\% as the training set, 40\% as the validation set, and the remaining 40\% as the test set. Then we compare our GraphAttacker with several attack methods on node classification. Our parameter analysis, average node similarity analysis, and attack performance are introduced as follows.

\begin{figure*}[htbp]\setlength{\belowcaptionskip}{-0.5cm}\setlength{\abovecaptionskip}{0.2cm}\vspace{-2em}
  \centering
  \includegraphics[width=0.85\linewidth]{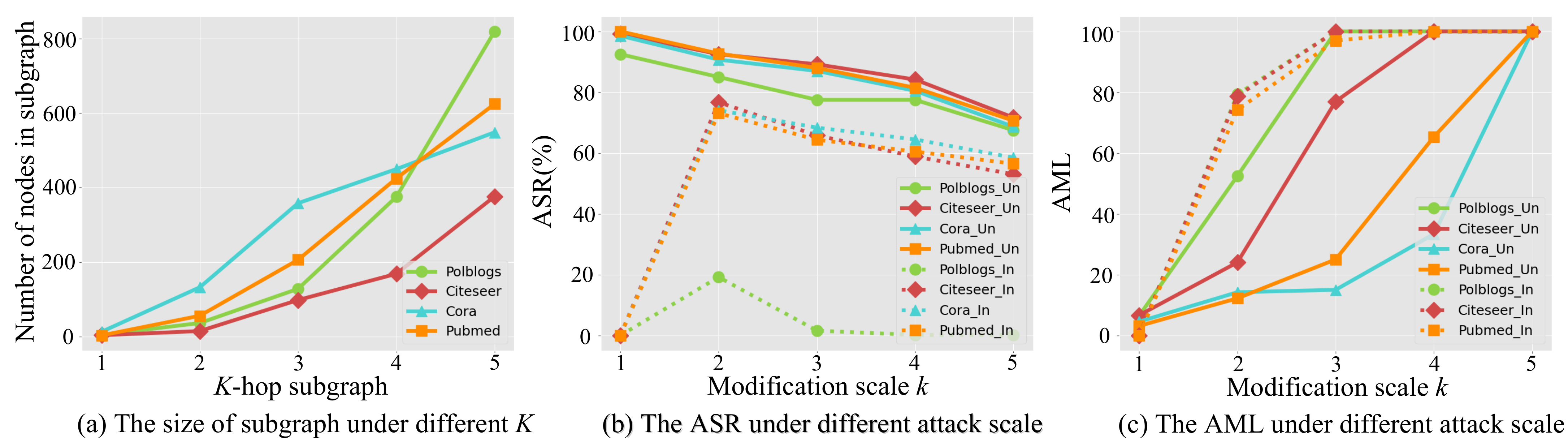}\\
  \caption{The subgraph size under different $K$ and the ASR /AML under different modification scales $k$. The solid/dotted line denotes unlimited/indirect attack. When $k=0$, we only modify the links or node attributes of the target node, which is a direct attack.}\label{fig.4}
\end{figure*}

\setlength{\parskip}{0\baselineskip}
\noindent\textbf{Parameter analysis.}  In node classification attack, $K$-hop subgraph is used  in the attack process to achieve an efficient and low-cost attack. Here, we mainly examine the impact of the chosen number of selected hops $K$, attack scale $k$ and the iteration times for MAG, SD, and AD on the performance of GraphAttacker.

\setlength{\parskip}{0\baselineskip}
We select the number of subgraph hops $K\in N_+$ to get proper $G_{K-sub}$ . As illustrated in {Fig.\ref{fig.4}(a), when $K$ is less than 3, the subgraph of the Cora, Citeseer, and Pubmed datasets only contain less than 50 nodes. Especially on the large-scale Pubmed dataset, the number of nodes in the $G_{K-sub}$ is even lower than 0.5\% of the original number of nodes.} In this case, we consider the information contained in the subgraph is insufficient. As $K$ increases, the size of the subgraph also increases rapidly. When $K=5$, {the subgraph size} has reached more than 10\% of the original graph. To achieve a more efficient attack, we set the maximum number of subgraph hops as $K=5$.

We further combine different attack scales $k$ to observe the attack effect. We choose $K=3$ as the initial setting of three attack scales, i.e. direct attack, indirect attack, and unlimited attack. Here we constrain the attack budget $\triangle \textless0.05 *|E|$ to get ASR. When the number of perturbations is greater than 100, AML is set to 100. When $K$ increases, we will also increase $k$ to perform the corresponding scale attacks. {Fig.\ref{fig.4}(b)} shows the ASR under different attack scales. The solid lines represent the unlimited attack results, and the dotted lines indicate the results of {the} indirect attack.
Since the initial value of $K$ is 3, {we are conducting a direct attack} when k = 0. At this time, we set the ASR and AML of indirect attack to 0.

We can see that our unlimited attack has better results than indirect attack. When $k=1$, i.e. performing a direct attack, the highest ASR and the lowest AML are obtained. Interestingly, with the increase of the scale $k$, ASR is decreased significantly in {Fig.\ref{fig.4}(b), while in contrast,} AML is increased in {Fig.\ref{fig.4}(c)}. We consider that because in $G_{K-sub}$, the effect of indirect links on the target node is different from that in $G$. Modifying the indirect links in the $G_{K-sub}$ can misclassify the target node. {Still, the impact of these perturbations may not be effective in $G$.}

\begin{figure}[htbp]\setlength{\belowcaptionskip}{-0.2cm}
  \centering
  \includegraphics[width=1\linewidth]{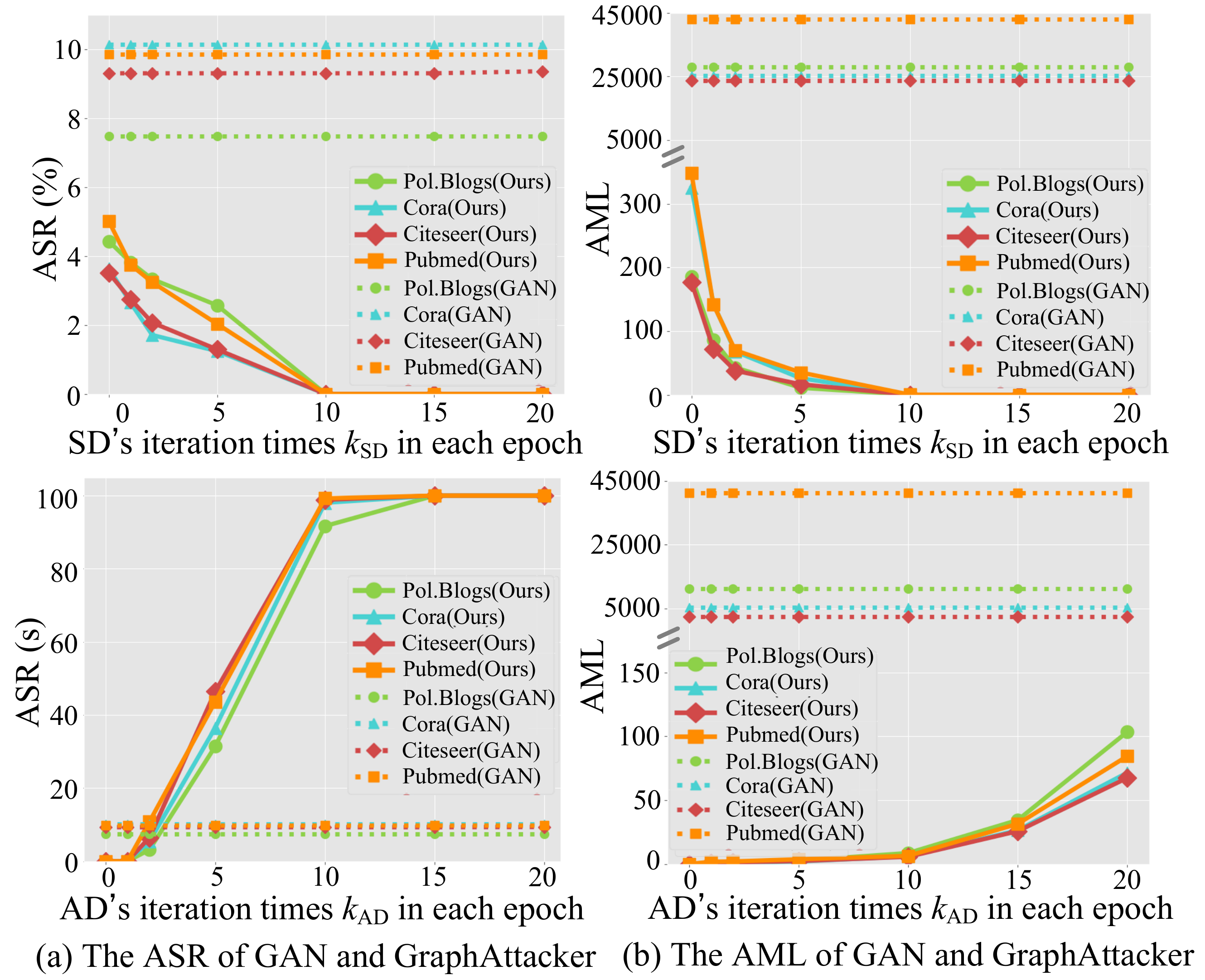}\\
  \caption{The solid line denotes the attack performance of GraphAttacker(Ours) under different $k_{SD}$ or $k_{AD}$, and the dashed line represents the attack performance of ``native" GAN.}\label{fig.GAN}
\end{figure}

\begin{figure*}[htbp]
\setlength{\belowcaptionskip}{-0.5cm}\setlength{\abovecaptionskip}{0.1cm}
  \centering
  \includegraphics[width=0.96\linewidth]{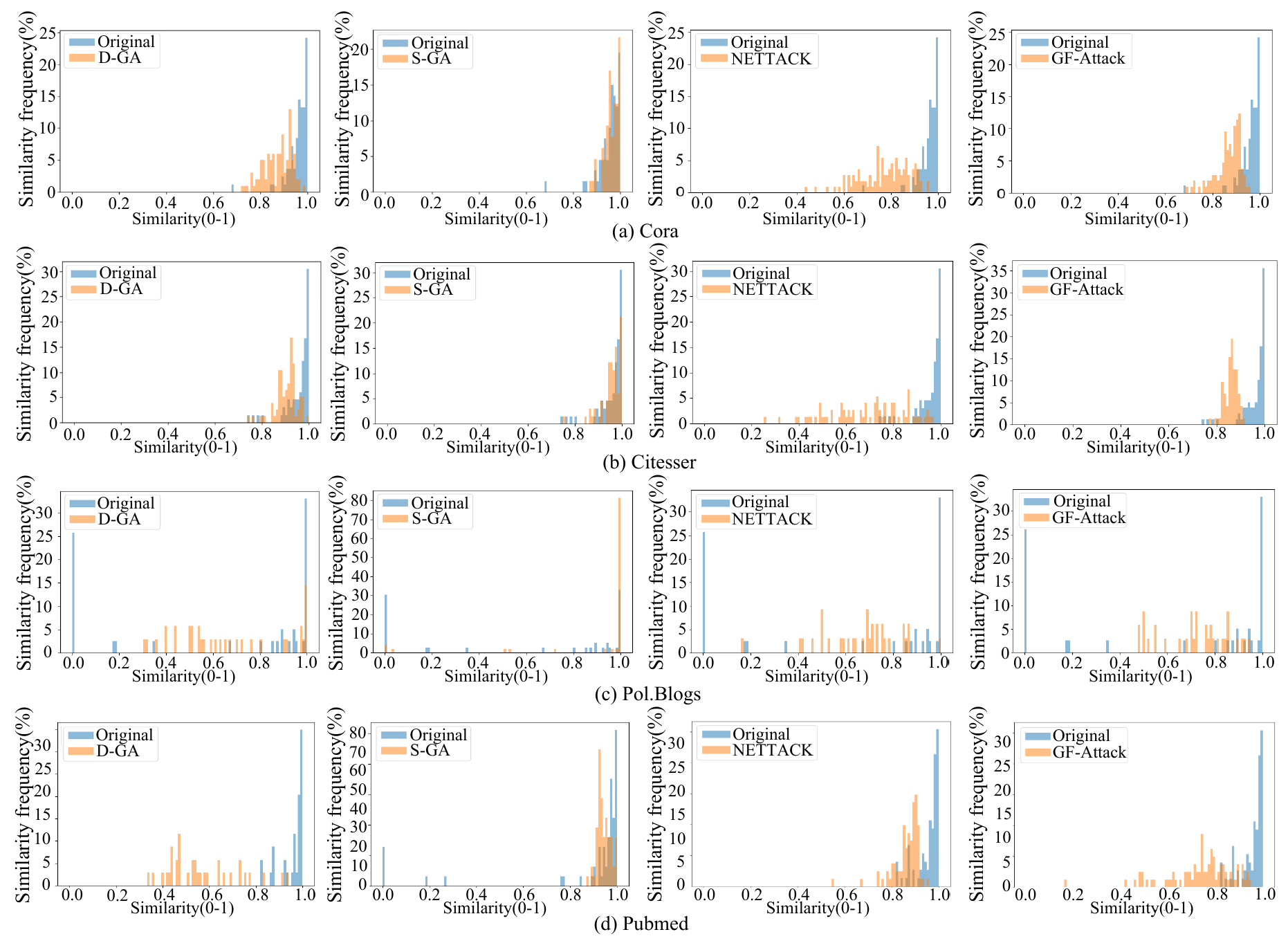}\\
  \caption{{Node similarity distribution on different datasets illustrates the average node similarity distribution change by different attack methods.  (D-GA: Degree-limited GraphAttacker; S-GA: Similarity-limited GraphAttacker)}}\label{fig.5}
\end{figure*}

As the parameters that control the three-player game of GraphAttacker, the selection of the number of iterations of MAG, SD, and AD in each epoch will directly determine the attack performance of GraphAttacker. Since GraphAttacker is designed based on GAN, we compare GraphAttacker with ``native" GAN to better illustrate the role of MAG, SD, and AD. Specifically, we use the source code released by the "native" GAN\cite{goodfellow2014generative} and its suggested parameter settings to generate adversarial examples with continuous-value. Then we discretize the adversarial examples through Eq.\ref{5}. In GraphAttacker, the number of subgraph hops $K$ and the attack scale $k$ are set to 3 and 1, respectively. The number of MAG¡¯s iterations $k_{MAG}$  is set to 10.

From Fig.\ref{fig.GAN}, the ``native" GAN can only achieve an ASR of less than 10\% under extremely high AML, which indicates that it is unsuitable for generating adversarial graph examples with excellent attack performance similar to the original ones. For GraphAttacker, we first set the number of AD's iterations $k_{AD}=0$ to explore the role of SD on GraphAttacker. As shown in Fig.\ref{fig.GAN} (a) and (b), as the number of SD's iterations $k_{SD}$ increase, GraphAttacker undergoes a decline in both ASR and AML gradually. This shows that SD can effectively guide MAG to generate adversarial examples similar to the original ones. Then, we set $k_{SD}=10$ and adjust the $k_{AD}$. Fig.\ref{fig.GAN} (c) and (d) indicate that AD effectively improves GraphAttacker's attack performance. When $k_{AD}=10$, GraphAttacker performs a satisfactory attack performance under the AML less than 10. As $k_{AD}$ continues to increase, although GraphAttacker's ASR has been slightly improved on Pol.Blogs, the price of the sharp increase in AML cannot be ignored.

\begin{table*}[]
\centering
\caption{The ASR and AML (mean and 95\% confidence interval) obtained by different attack methods on node classification tasks. We use \textbf{bold} to highlight wins. (B-GA:Budget-limited GraphAttacker; D-GA:Budget-limited GraphAttacker; S-GA:Similarity-limited GraphAttacker )}
\resizebox{185mm}{43mm}{
\begin{tabular}{cccccc|cccccc}
\hline
\multicolumn{1}{l}{\multirow{3}{*}{Metric}} & \multicolumn{1}{l}{\multirow{3}{*}{Dataset}}    & \multicolumn{1}{l}{\multirow{3}{*}{Model}} & \multicolumn{9}{c}{Attack methods}                                                                                                                                                                                                                   \\ \cline{4-12}
\multicolumn{1}{l}{}                        & \multicolumn{1}{l}{}                            & \multicolumn{1}{l}{}                       & \multicolumn{3}{c|}{Ours}                                                                                       & \multicolumn{6}{c}{Baselines}                                                                                                     \\ \cline{4-12}
                        &                            & \multicolumn{1}{c}{}                       & B-GA                         & D-GA                         & S-GA                         & DICE                          & NETTACK                       & GF-Attack      & GUA             & IG-JSMA        & SGA            \\ \hline
\multirow{12}{*}{ASR(\%)}   & \multirow{3}{*}{Pol.Blogs} & GCN                                         & \textbf{89.99$\pm$1.40}             & 83.33$\pm$1.86               & 51.28$\pm$2.85               & 50.27$\pm$4.43                & 82.97$\pm$1.74                & 19.89$\pm$3.41 & 46.42$\pm$3.41  & 89.14$\pm$1.74 & 86.68$\pm$1.34 \\
                        &                            & Deepwalk                                    & \textbf{87.45$\pm$1.86}               & 79.49$\pm$1.94               & 47.41$\pm$2.45               & 64.52$\pm$3.76                & 75.41$\pm$1.83                & 12.82$\pm$3.26 & 34.14$\pm$5.47 & 85.47$\pm$2.67 & 86.04$\pm$1.62 \\
                        &                            & LINE                                        & \textbf{88.47$\pm$2.06}               & 81.74$\pm$2.01               & 48.67$\pm$1.33               & 66.74$\pm$4.03                & 76.35$\pm$1.88                & 23.48$\pm$3.54 & 37.41$\pm$4.70  & 88.13$\pm$1.78 & 86.44$\pm$1.85 \\ \cline{2-12}
                        & \multirow{3}{*}{Cora}      & GCN                                         & \textbf{98.41$\pm$1.59}              & 92.43$\pm$1.03               & 78.26$\pm$2.23               & 52.36$\pm$2.49                & 91.04$\pm$1.83                & 80.67$\pm$1.88 & 84.68$\pm$3.08  & 97.64$\pm$1.56 & 94.74$\pm$2.14 \\
                        &                            & Deepwalk                                    & \textbf{96.02$\pm$1.09}             & 91.47$\pm$1.34               & 78.61$\pm$2.16               & 91.84$\pm$2.84                & 92.64$\pm$1.54                & 59.43$\pm$3.45 & 69.41$\pm$6.14  & 94.74$\pm$2.67 & 94.06$\pm$2.42 \\
                        &                            & LINE                                        & \textbf{96.75$\pm$1.33}               & 90.54$\pm$1.60               & 77.25$\pm$2.66               & 85.47$\pm$3.43                & 94.44$\pm$2.13                & 80.14$\pm$2.34 & 72.41$\pm$5.47  & 94.74$\pm$2.14 & 95.04$\pm$2.33 \\ \cline{2-12}
                        & \multirow{3}{*}{Citeseer}  & GCN                                         & \textbf{98.53$\pm$1.47}               & 95.87$\pm$1.63               & 81.12$\pm$2.34               & 67.14$\pm$3.59                & 86.01$\pm$1.65                & 58.47$\pm$2.89 & 78.16$\pm$4.13  & 97.16$\pm$1.57 & 92.79$\pm$2.26 \\
                        &                            & Deepwalk                                    & \textbf{97.14$\pm$1.24}               & 95.84$\pm$1.87               & 80.87$\pm$2.04               & 90.47$\pm$3.01                & 95.13$\pm$1.57                & 48.41$\pm$2.34 & 64.14$\pm$4.87  & 97.10$\pm$1.41 & 96.26$\pm$1.97 \\
                        &                            & LINE                                        & \textbf{97.49$\pm$2.51}               & 94.74$\pm$1.87               & 81.26$\pm$2.26               & 92.43$\pm$4.04                & 93.96$\pm$1.76                & 58.24$\pm$2.14 & 64.75$\pm$5.62  & 97.23$\pm$1.51 & 95.86$\pm$2.06 \\ \cline{2-12}
                        & \multirow{3}{*}{Pumbed}    & GCN                                         & \textbf{98.62$\pm$1.27}               & 97.41$\pm$2.59               & 66.41$\pm$2.16               & 51.44$\pm$3.63                & 82.14$\pm$1.54                & 52.47$\pm$2.34 & 85.45$\pm$3.47  & 98.33$\pm$1.41 & 95.43$\pm$2.04 \\
                        &                            & Deepwalk                                    & \textbf{97.74$\pm$2.26}               & 94.13$\pm$2.43               & 66.14$\pm$2.30               & 51.04$\pm$3.47                & 82.14$\pm$1.78                & 44.47$\pm$2.04 & 77.47$\pm$4.56  & 97.26$\pm$2.41 & 95.63$\pm$2.66 \\
                        &                            & LINE                                        & \textbf{97.16$\pm$1.75}               & 94.74$\pm$2.13               & 63.47$\pm$3.13               & 59.74$\pm$3.64                & 91.72$\pm$1.74                & 45.74$\pm$2.67 & 75.47$\pm$5.34  & 96.54$\pm$2.16 & 96.42$\pm$2.40 \\ \hline
\multirow{12}{*}{AML}   & \multirow{3}{*}{Pol.Blogs} & GCN                                         & 8.62$\pm$0.48                & 7.25$\pm$0.86                & \textbf{6.38$\pm$1.43}      & 11.85$\pm$2.57        & 11.89$\pm$0.76         & 20.00$\pm$0.00 & 20.00$\pm$0.00  & 6.42$\pm$1.47  & 6.87$\pm$1.71  \\
                        &        & Deepwalk      & 9.15$\pm$0.87 & 7.41$\pm$1.14 & 6.96$\pm$1.39 & 12.35$\pm$2.74 & 10.06$\pm$0.84 & 20.00$\pm$0.00 & 20.00$\pm$0.00  & \textbf{6.24$\pm$1.41} & 7.84$\pm$1.67  \\
                        &                            & LINE                                        & 8.83$\pm$0.84                & 7.17$\pm$1.07                & 6.62$\pm$1.48                & 12.82$\pm$2.63                & 10.26$\pm$1.14                & 20.00$\pm$0.00 & 20.00$\pm$0.00  & \textbf{6.07$\pm$1.13} & 7.35$\pm$1.24  \\ \cline{2-12}
                        & \multirow{3}{*}{Cora}      & GCN                                         & 6.09$\pm$0.58                & 5.47$\pm$1.07                & \textbf{4.39$\pm$1.73}                & 9.13$\pm$2.47                 & 6.09$\pm$0.87                 & 20.00$\pm$0.00 & 7.52$\pm$3.83   & 4.79$\pm$0.47  & 5.61$\pm$0.77  \\
                        &                            & Deepwalk                                    & 6.47$\pm$0.88                & 5.41$\pm$1.16                & 5.12$\pm$1.54                & 7.20$\pm$2.85                 & 7.24$\pm$0.92                 & 20.00$\pm$0.00 & 8.17$\pm$3.47   & \textbf{5.04$\pm$0.73}  & 5.76$\pm$1.41  \\
                        &                            & LINE                                        & 6.52$\pm$0.82                & 5.66$\pm$1.07                & \textbf{5.34$\pm$1.67}                & 7.66$\pm$2.46                 & 7.02$\pm$1.35                 & 20.00$\pm$0.00 & 7.82$\pm$3.19   & 5.47$\pm$0.76  & 6.04$\pm$1.29  \\ \cline{2-12}
                        & \multirow{3}{*}{Citeseer}  & GCN                                         & 5.71$\pm$0.75                & 4.04$\pm$1.13                & 4.29$\pm$1.36                & 9.87$\pm$2.41                 & 6.88$\pm$0.87                 & 20.00$\pm$0.00 & 11.45$\pm$2.87  & \textbf{3.85$\pm$0.47}  & 4.16$\pm$0.88  \\
                        &                            & Deepwalk                                    & 6.46$\pm$1.04                & 4.95$\pm$1.57                & 4.89$\pm$1.35                & 7.08$\pm$2.68                 & 7.06$\pm$1.28                 & 20.00$\pm$0.00 & 10.86$\pm$2.75  & \textbf{4.15$\pm$0.86} & 4.82$\pm$1.20  \\
                        &                            & LINE                                        & 6.57$\pm$1.08                & 4.99$\pm$1.55                & 5.01$\pm$1.44                & 7.21$\pm$1.78                 & 6.02$\pm$1.30                 & 20.00$\pm$0.00 & 11.03$\pm$3.03  &\textbf{ 4.32$\pm$0.84 } & 5.14$\pm$1.27  \\ \cline{2-12}
                        & \multirow{3}{*}{Pumbed}    & GCN                                         & 6.42$\pm$0.69                & 5.67$\pm$0.77                & \textbf{5.11$\pm$1.13}                & 10.30$\pm$3.14                & 8.14$\pm$0.88                 & 20.00$\pm$0.00 & 18.43$\pm$3.87  & 5.23$\pm$0.61  & 4.52$\pm$0.98  \\
                        &                            & Deepwalk                                    & 7.03$\pm$0.87                & 5.94$\pm$1.43                & 5.74$\pm$1.24                & 10.87$\pm$3.04                & 7.66$\pm$0.94                 & 20.00$\pm$0.00 & 18.84$\pm$4.14  & \textbf{5.33$\pm$0.86} & 5.12$\pm$1.04  \\
                        &                            & LINE                                        & 7.21$\pm$1.34                & 6.01$\pm$1.39                & 5.80$\pm$1.18                & 10.13$\pm$3.61                & 8.05$\pm$1.14                 & 20.00$\pm$0.00 & 19.14$\pm$3.96  & \textbf{5.74$\pm$1.03}  & 5.34$\pm$1.41  \\ \hline
\end{tabular}}\label{tab:2}
\end{table*}

\setlength{\parskip}{0\baselineskip}
\noindent\textbf{Attack performance.} We refer to our GraphAttacker under the attack budget $\triangle \textless 0.05*|E|$ as B-GA, and the GraphAttacker under  $\triangle \textless 0.05*|E|$ and $\Lambda\textless 0.004$ as D-GA. The node similarity modification ratio constraint is realized by adding high similarity nodes to $G_{K-sub}$ with the constraint as $\triangle \textless 0.05*|E|$, $\Lambda\textless 0.004$, and $SMR\textless 0.05$, which we call S-GA. {According to the experimental results above, we set $K=3$ in GraphAttacker, using graph structure attack and direct attack}, i.e., attack $A$ and set $k=0$ to obtain ASR and AML, and compare the attack results with several other attack methods in {TABLE~\ref{tab:2}.}

 \setlength{\parskip}{0\baselineskip}
 We can observe that B-GA outperforms the baseline attack methods with higher ASR in all the cases. For the D-GA under the constraints of degree distribution statistics $\Lambda$, IG-JSMA and SGA are two powerful competitors. IG-JSMA searches for the perturbations that have the greatest impact on the model's prediction results according to the target model's integral gradient, which makes it achieve effective attacks with the lowest AML. However, the better performance of B-GA indicates that there may still be slight errors in the integral gradient, although it has largely solved the issue of inaccurate gradients on the discrete data. As another gradient-based attack method, SGA utilizes the target nodes' degree as the attack budget $\triangle$ and performs similar attack performance to D-GA in terms of ASR and AML. Additionally, when attacking the random walk-based graph embedding methods, GraphAttacker also has a higher ASR, which proves its strong transferability.

 An interesting result is that attacks under higher stealthiness constraints, such as D-GA and S-GA, have lower AML than B-GA. The reason is that in the iterative optimization process of GraphAttacker, once a generated adversarial example can successfully achieve the attack under the given constraints, GraphAttacker will stop the attack. Therefore, although the adversarial examples generated by GraphAttacker can meet our attack requirements, they are usually not optimal. Higher stealthiness constraints will increase the cost of attacks. However, once the attack is successful, it can bring the benefits of lower AML.

\begin{table*}[htbp]
\centering
\caption{The ASR, AML and AMA (mean and 95\% confidence interval) obtained by GraphAttacker and GraphAttacker-ori with different attack strategies. We use \textbf{bold} to highlight wins. }
\resizebox{178mm}{20mm}{
\begin{tabular}{ccccc|cc|cc}
\hline
\multirow{2}{*}{Attack method}     & \multirow{2}{*}{Dataset} & \multicolumn{3}{c|}{ASR(\%)}   & \multicolumn{2}{c|}{AML}       & \multicolumn{2}{c}{AMA}         \\ \cline{3-9}
                                   &                          & A     & X     & Hybrid         & A             & Hybrid         & X              & Hybrid         \\ \hline
\multirow{4}{*}{GraphAttacker}     & Pol.Blogs                & 89.99$\pm$1.40  & 65.30$\pm$2.20  & \textbf{93.84$\pm$1.16}    & 8.62$\pm$0.84         & \textbf{5.43$\pm$0.92}  & 5.63$\pm$0.66          & \textbf{4.30$\pm$0.74}   \\
                                   & Cora                     & {98.41$\pm$1.59} & 49.29$\pm$2.14 & \textbf{98.64$\pm$1.36} & 6.09$\pm$1.06         & \textbf{5.79$\pm$1.14}  & \textbf{12.88$\pm$3.41} & 13.93$\pm$3.88          \\
                                   & Citeseer                 & {98.53$\pm$1.47} & 61.67$\pm$1.87 & \textbf{98.58$\pm$1.42}   & 5.71$\pm$0.97         & \textbf{3.06$\pm$0.85}  & \textbf{33.48$\pm$4.47} & 34.97$\pm$4.19   \\
                                   & {Pubmed }                & {{98.62$\pm$1.27}} & {63.42$\pm$1.92} & {\textbf{98.84$\pm$1.16}}   & {6.42$\pm$1.24 }        & {\textbf{4.41$\pm$0.92}}  & {19.42$\pm$2.04} & {\textbf{18.64$\pm$1.86}}   \\ \hline
\multirow{4}{*}{GraphAttacker-ori} & Pol.Blogs                & 65.14$\pm$2.36  & 45.10$\pm$2.40  & \textbf{70.61$\pm$1.89}  & 18.23$\pm$2.87         & \textbf{17.12$\pm$2.44} & \textbf{9.14$\pm$2.13}  & 10.53$\pm$1.96          \\
                                   & Cora                     & 64.73$\pm$1.70 & 27.86$\pm$2.11 & \textbf{68.01$\pm$1.28}    & \textbf{3.58$\pm$0.48} & 3.72$\pm$0.53           & 7.48$\pm$1.78            & \textbf{7.04$\pm$1.74}  \\
                                   & Citeseer                 & 68.47$\pm$0.70 & 45.83$\pm$3.47 & \textbf{69.29$\pm$2.47}  & 5.52$\pm$1.04         & \textbf{4.97$\pm$0.88}  & 21.97$\pm$3.10          & \textbf{19.47$\pm$2.59} \\
                                   & {Pubmed}                 & {64.67$\pm$2.00} & {46.12$\pm$2.88} & {\textbf{66.15$\pm$2.18}}   & {6.42$\pm$1.27}         & {\textbf{4.41$\pm$1.05}}  &{ 19.42$\pm$2.57} & {\textbf{18.64$\pm$2.27}}   \\\hline
\end{tabular}
\label{tab:3}%
}
\end{table*}

 Due to stronger stealthiness constraints, S-GA has {a} smaller ASR than B-GA and D-GA. In the dataset Pol.Blog, the ASR of S-GA dropped significantly. That is because the nodes in Pol.Blogs have obvious differentiation. {Only a few nodes have high similarity with other categories of nodes, making} it difficult for GraphAttacker to generate effective attack perturbations in the subgraph. To verify whether S-GA sacrifices the ASR for higher stealthiness, we plot the average similarity distribution of the target node before and after the attack with different attack methods in {Fig.\ref{fig.5}}, where the x-axis denotes the {average similarity (AS)} value between the target node and its linked nodes. {The y-axis is the frequency of nodes with different average similarities. D-GA}, NETTACK, and GF-Attack have significantly changed the average similarity value of the target nodes. In particular, in  Pol.Blogs, the AS of the nodes is roughly concentrated at $0$ or $1$, and the nodes with $AS=0$ are isolated nodes. {Under S-GA attack, these isolated nodes are connected with highly similar nodes to be better hidden in normal nodes.} Combining {TABLE~\ref{tab:2}} and {Fig.\ref{fig.5}}, although S-GA sacrifices part of ASR, it generally maintains the original average similarity value of the target node, which is successfully attacked under the constraint of $\triangle$, $\Lambda$, and $SMR$. We believe that our GraphAttacker can further {enhance} the stealthiness of the perturbations.

\begin{figure}
\setlength{\belowcaptionskip}{-0.6cm}\setlength{\abovecaptionskip}{0.1cm}\vspace{-1em}
  \centering
  \includegraphics[width=0.7\linewidth]{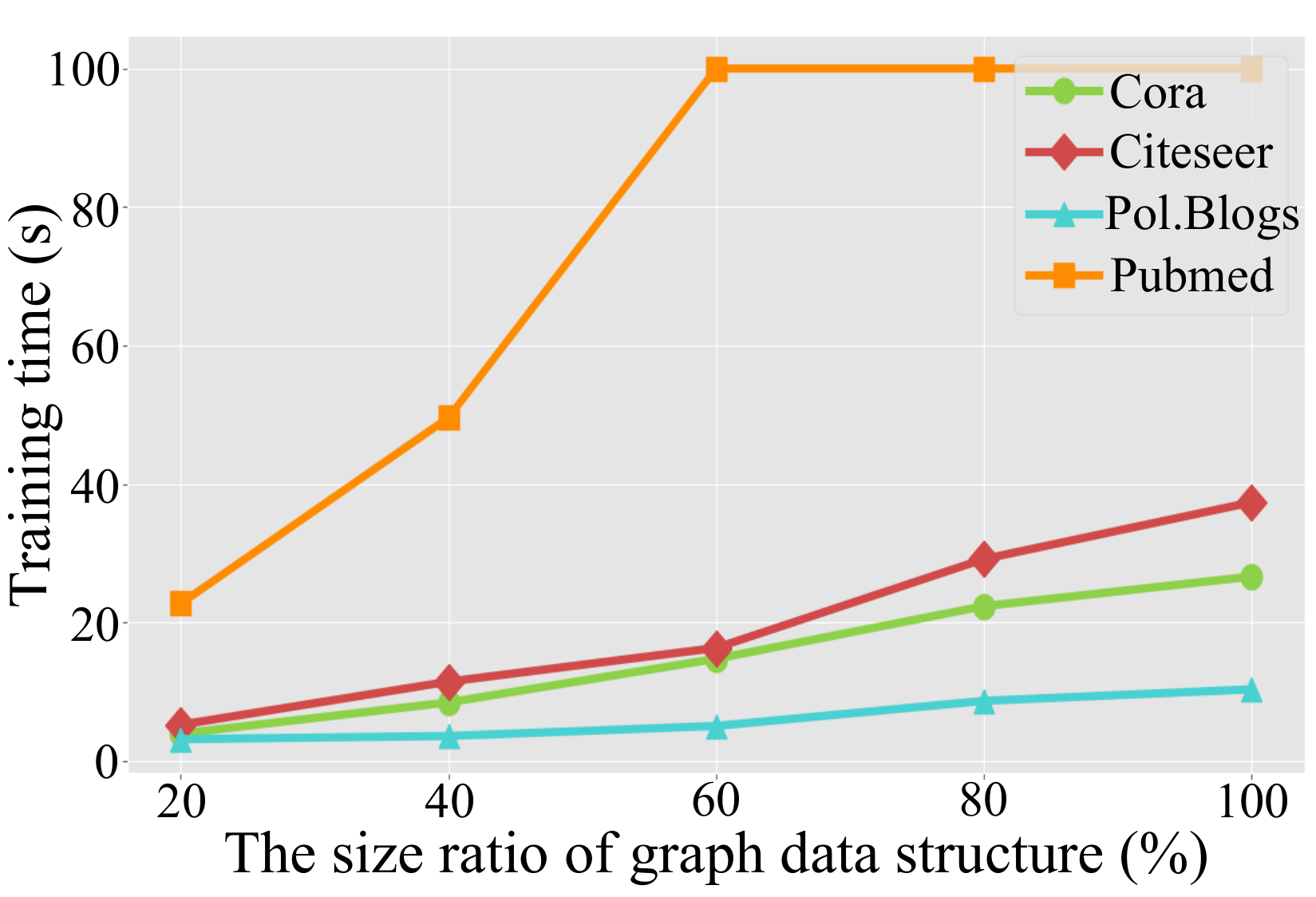}\\
  \caption{{The training time in each iteration when attacking different size of graph structure.}}
  \label{fig:6}
\end{figure}
\vspace{0cm}

\setlength{\parskip}{0\baselineskip}
\noindent\textbf{Different modification strategies.} We also implement node attribute attack and hybrid attack and compare them with the GraphAttacker-ori without adding random nodes to the $G_{K-sub}$. Here we also set $K$ to $3$ and use {the} direct attack. When attacking node attributes, we set the stealthiness constraint as $\triangle \textless 0.05*|E|$. In each epoch of the hybrid attack, the MAG generates the adversarial subgraph adjacency matrix $A_{sub}^{'}$ , and then generates the adversarial node attributes $X_{sub}^{'}$ based on $A_{sub}^{'}$. From {TABLE~\ref{tab:3}}, we can find that the attack on node attributes can only obtain less than $50\%$ ASR. This may be that the node attributes of the citation datasets are relatively sparse, and the node categories are more determined by the graph structure. In a hybrid attack, we can get the highest ASR while reducing the AML by modifying the node attributes. Similar results can be observed in GraphAttacker-ori. {However,} the ASR of GraphAttacker-ori is much lower than that of GraphAttacker. It confirms our conjecture that the category distribution of the nodes is relatively concentrated in the original neighbor subgraph of the target node, {which makes the attack very difficult. The strategy of randomly adding nodes to the subgraph is effective.}

\setlength{\parskip}{0\baselineskip}
\noindent\textbf{Time efficiency of attack.} Most existing methods iteratively generate adversarial perturbations step by step. Different from them, GraphAttacker directly generates adversarial examples by its generator. In {Fig.\ref{fig:6}, we show the training time of GraphAttacker for different sizes of data structures. When the training time of each iteration is more than 100s, we set it to 100s}. {As the size of the graph structure increases, the training time also increases significantly. Since our $K$-hop neighbor subgraph size is smaller than 20\% of the original graph, our strategy of attacking the $K$-hop neighbor subgraph can effectively reduce the attack cost. Especially for large-scale datasets such as Pubmed, GraphAttacker can effectively avoid the high complexity caused by GAN.}

\setlength{\parskip}{0\baselineskip}\label{graphatt}
\subsubsection{Graph classification attack} {In the previous part,} we introduce the node similarity stealthiness constraints and attack performance of GraphAttacker on the node classification task. In this part,
we mainly explore the attack performance of GraphAttacker on graph classification {tasks} under different modification strategies.

\setlength{\parskip}{0\baselineskip}
\noindent\textbf{Model configurations.} {Following several previous works \cite{ying2018hierarchical,lee2019self}, we randomly
split each graph classification dataset into three parts: 80\% as the training set, 10\% as the validation set, and the remaining 10\% as the test set. We use Diffpool\cite{ying2018hierarchical} built on the GCN architecture as our target attack model. The structure and parameter settings of Diffpool are consistent with the source code released by the author. It constructs two embedding GCNs and one pooling GCN by using a three-layer GCN module. According to the original graph, the first embedding GCN and the pooling GCN generate a node embedding and an assignment matrix, respectively.} Then use the original graph, node embedding, and assignment matrix to generate a coarsened graph and node attributes, and use them as the input of the second embedding GCN to get a new node embedding. Finally, a fully connected layer is used to obtain the classification result of the target graph.

\setlength{\parskip}{0\baselineskip}
\noindent\textbf{Attack strategy.} From the dataset statistics in {TABLE~\ref{tab:1}}, we can see that the datasets used for graph classification usually consist of many graphs. Still the average size of these graphs is much smaller than that of citations and social network datasets. Therefore, without considering the difficulty of GAN in generating a large-size graph, we use GraphAttacker to directly generate the whole adversarial examples. Since the graph classification task predicts the graph category by aggregating the information of all nodes in the graph, it pays more attention to the overall characteristics of the graph. Therefore, we don't need to consider direct attack and indirect attack in graph classification attack. Here we have conducted an unlimited attack. For modification strategy, similar to the node classification attack, we carry out graph structure attack, node attribute attack, and hybrid attack in  graph classification attack.


\noindent\textbf{Attack performance.} Here we show the attack performance of different modification strategies on graph classification attack.

\setlength{\parskip}{0\baselineskip}
\textbf{Structure attack.} We constrain the attack budget on the graph structure to $\triangle \textless r *|N|^{2}$, where $r$ is the perturbation ratio. The tendency of GraphAttacker's ASR with different structural perturbation ratios is shown in {Fig.\ref{fig:7} (a)}. On the one hand, as perturbations increases, the ASR on different datasets also increases. On the other hand, when the max ratio of structural
perturbation is less than $10\%$, the ASR increases rapidly. After that, ASR does not increase significantly. It indicates that in graph classification attack, blindly increasing the attack budget may not achieve higher ASR. {We speculate that in those datasets with low ASR, the graph structure may not focus on the graph classification model.}

\begin{figure}[htb]
\setlength{\belowcaptionskip}{-0.5cm}\setlength{\abovecaptionskip}{0.2cm}\vspace{-1em}
  \centering
  \includegraphics[width=1\linewidth]{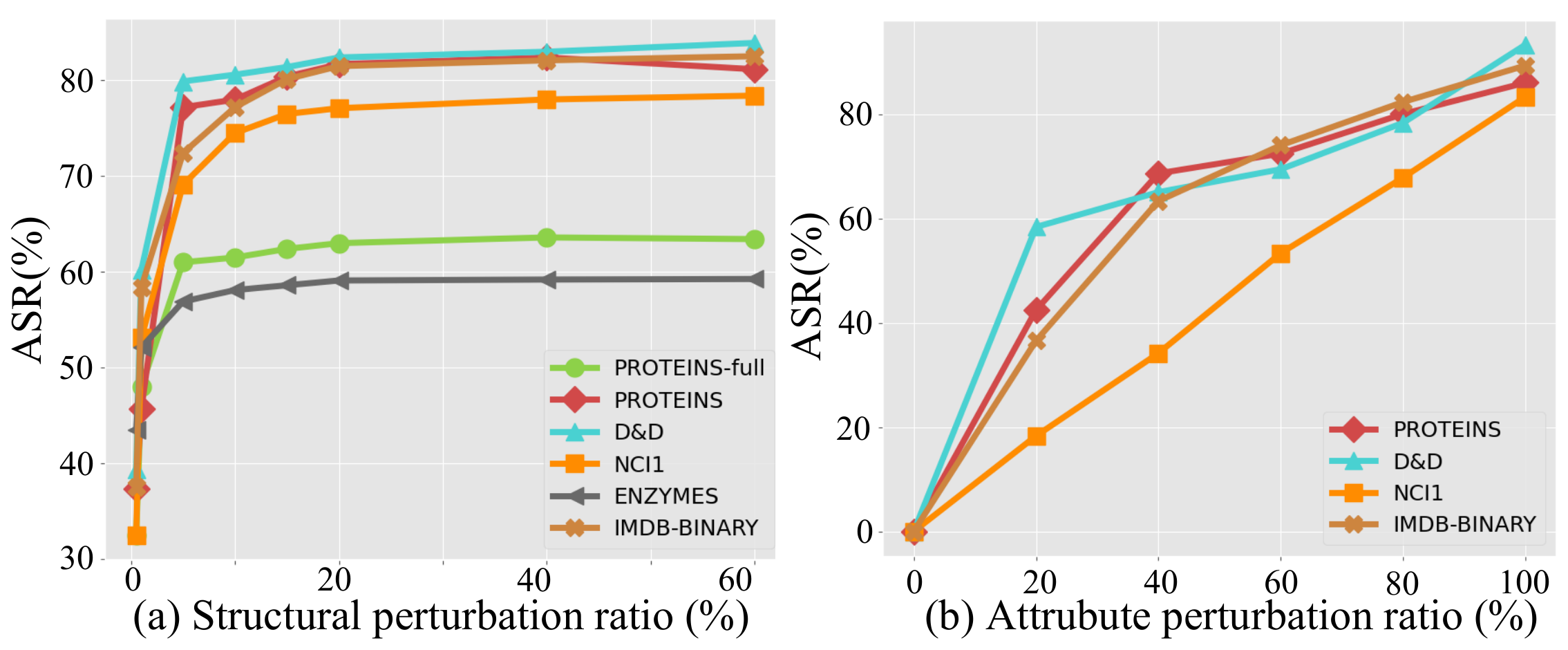}\\
  \caption{ASR of different structural and attribute perturbation ratios.}
  \label{fig:7}
\end{figure}

\textbf{Attribute and hybrid attacks.} According to the types of node attributes, the datasets used for graph analysis are divided into two types. {One uses node category as one-dimensional node attributes, such as PROTEIN, DD, NCI1, and IMDB-BINARY.} The other has multi-dimensional continuous node attributes, such as PROTEIN-full and ENZYMES.

\setlength{\parskip}{0\baselineskip}

For the first type of datasets, we use one-hot encoding to express their node attributes: $X \in\{0,1\}^{N \times |F|}$, where $F$ denotes the number of categories of nodes, and each node only has one attribute. We can still use $\triangle$ to constrain the perturbations generated by GraphAttacker on the node attributes for these datasets. {Fig.\ref{fig:7}(b)} shows the ASR of the first type of datasets under different attribute perturbation ratios $r$, and the attack budget is set as: $\triangle\textless r*|N|$. Comparing with the highest values of ASR in {Fig.\ref{fig:7}(a)} and {Fig.\ref{fig:7}(b)}, node attribute attack can achieve higher ASR. However, it requires a larger attack budget. For example, on PROTEIN, the ASR can reach $51.35\%$ under $1\%$ structural perturbation ratio constraint, which is higher than the ASR($42.49\%$) under the $20\%$ node attribute perturbation ratio constraint.
Because the node attributes composed of node categories are much more sparse and contain much less information than the graph structure, it is more difficult to influence the prediction results for the target model.

For the second type of datasets, {since the $\triangle$ and $\Lambda$ do not apply to continuous node attributes}, we use L2-Norm to measure the perturbation size of continuous node attributes and use it as an additional optimization item for AD.

To better understand the difficulty of attack graph structure or node attributes, we compare the attack effects of different modification strategies (including hybrid attack) under the constraint of $r\textless20\%$. As shown in {TABLE~\ref{tab:4}}, the hybrid attack achieves the highest ASR on all datasets, which has lower AML or L2-Norm in most cases. For the first type of datasets, their graph structures are relatively dense. Therefore, graph structure attack can achieve higher ASR than node attribute attack under the same constraints. {For the second type of datasets, the continuous node attributes are denser, and the above results are the opposite.}

\begin{table*}[htpb]
\centering
\caption{Attack performance of graph classification tasks under different modification strategies.}
\resizebox{178mm}{14mm}{
\begin{tabular}{cccc|cc|cc|cc}
\hline
\multirow{2}{*}{Dataset} & \multicolumn{3}{c|}{ASR(\%)} & \multicolumn{2}{c|}{AML} & \multicolumn{2}{c|}{AMA} & \multicolumn{2}{c}{L2 Norm} \\ \cline{2-10}
                         & A       & X       & Hybrid  & A           & Hybrid     & X          & Hybrid      & X         & Hybrid      \\ \hline
PROTEINS-full            & 63.42$\pm$1.56   & 94.88$\pm$1.75   & \textbf{97.62$\pm$1.03}   & 21.32$\pm$3.31       & \textbf{14.87$\pm$2.72}      & -          & -           & 1.61$\pm$0.37      & \textbf{1.48$\pm$0.39}        \\
PROTEINS                 & 81.17$\pm$1.32   & 42.49$\pm$4.28   & \textbf{84.32$\pm$1.85}   & \textbf{22.30$\pm$3.47}        & 24.63$\pm$3.18      & 7.27$\pm$1.87       & \textbf{4.45$\pm$1.15}        & -         & -           \\
D\&D                     & 83.92$\pm$1.26   & 58.42$\pm$3.36   & \textbf{96.14$\pm$1.61}   & 678.74$\pm$34.07      & \textbf{612.96$\pm$29.47}     & 14.82$\pm$2.11      & \textbf{13.72$\pm$2.16}       & -         & -           \\
NCI1                     & 78.41$\pm$1.60   & 18.33$\pm$1.63   & \textbf{87.86$\pm$2.13}   & 78.79$\pm$5.14      & \textbf{31.80$\pm$2.29}       & \textbf{3.50$\pm$0.87}        & 4.42$\pm$1.12        & -         & -           \\
ENZYMES                  & 59.26$\pm$2.41   & 88.62$\pm$1.46   & \textbf{90.37$\pm$1.27}   & 20.82$\pm$2.52       & \textbf{16.05$\pm$2.26}      & -          & -           & 1.57$\pm$0.44      & \textbf{1.51$\pm$0.52}        \\
{IMDB-BINARY }             & {82.53$\pm$1.87}   & {36.60$\pm$2.36 }  & {\textbf{87.46$\pm$1.88}}  & {31.51$\pm$3.20 }       &{\textbf{ 20.24$\pm$2.41 }}     & {14.69$\pm$2.51}     & {\textbf{14.03$\pm$1.63}}        & -         & -  \\ \hline
\end{tabular}}
\label{tab:4}%
\end{table*}

\begin{table*}[htbp]
\centering
\caption{The ASR and AML (mean and 95\% confidence interval) obtained by different attack methods on the link prediction task. (B-GA:Budget-limited GraphAttacker; D-GA:Budget-limited GraphAttacker; S-GA:Similarity-limited GraphAttacker )}
\resizebox{178mm}{22mm}{
\begin{tabular}{ccccccc|ccc|cc}
\hline
\multirow{3}{*}{Dataset}  & \multirow{3}{*}{Model} & \multicolumn{5}{c|}{ASR}                                                                                                                & \multicolumn{5}{c}{AML}                                                                                                             \\ \cline{3-12}
                          &                        & \multicolumn{3}{c|}{Ours}                                                        & \multicolumn{2}{c|}{Baseline}                        & \multicolumn{3}{c|}{Ours}                                                       & \multicolumn{2}{c}{Baseline}                      \\ \cline{3-12}
                          &                        & \multicolumn{1}{c}{B-GA} & \multicolumn{1}{c}{D-GA} & \multicolumn{1}{c|}{S-GA}  & \multicolumn{1}{c}{IGA} & \multicolumn{1}{c|}{DICE} & \multicolumn{1}{c}{B-GA} & \multicolumn{1}{c}{D-GA} & \multicolumn{1}{c|}{S-GA} & \multicolumn{1}{c}{IGA} & \multicolumn{1}{c}{DICE} \\ \hline
                          \multirow{3}{*}{Cora}     & {}{GAE}     &{}{\textbf{64.95$\pm$2.84}}  & {}{58.76$\pm$2.55}     & \multicolumn{1}{l|}{{}{48.06$\pm$2.96}} & {}{61.42$\pm$1.27}    & {}{4.60$\pm$3.89}    &{}{53.57$\pm$4.57}	&{}{42.68$\pm$3.88}	&{}{\textbf{38.15$\pm$3.97}}	&{}{45.54$\pm$4.12}	&{}{53.12$\pm$4.39}

 \\
                          & {}{Deepwalk    }           & {}{\textbf{95.14$\pm$1.53} }          & {}{93.14$\pm$1.17}    & \multicolumn{1}{l|}{{}{85.84$\pm$1.24}} & {}{92.71$\pm$2.47}	&{}{61.90$\pm$1.69}	&{}{34.19$\pm$2.34}	&{}{29.41$\pm$3.44}	&{}{\textbf{21.26$\pm$4.75}}	&{}{33.68$\pm$4.57}	&{}{23.37$\pm$5.86}
 \\
                          & {Node2vec}               & {}{\textbf{95.24$\pm$1.46} }          & {}{93.26$\pm$1.88 }                  & \multicolumn{1}{l|}{{}{84.12$\pm$2.21}} & {}{92.57$\pm$1.67}	& {}{58.67$\pm$2.89} & 	{35.62$\pm$3.37}	& {}{29.84$\pm$3.84}	& {}{\textbf{20.47$\pm$4.10}}	& {}{20.84$\pm$3.76}	& {24.51$\pm$5.33}  \\ \hline

\multirow{3}{*}{NS}       & GAE                    & \textbf{56.25$\pm$1.56}           & 44.33$\pm$2.98                    & \multicolumn{1}{l|}{40.21$\pm$2.14} & 56.20$\pm$2.47                  & 1.82$\pm$0.86                    & 21.31$\pm$3.22                    & 14.83$\pm$2.87                    & \textbf{5.24$\pm$1.24}             & 8.04$\pm$1.58                    & 11.29$\pm$2.03                    \\
                          & Deepwalk               & \textbf{96.47$\pm$2.27}           & 89.69$\pm$2.30                    & \multicolumn{1}{l|}{85.57$\pm$2.87} & 76.83$\pm$2.45                   & 49.81$\pm$1.85                    & 11.23$\pm$2.35                    & 6.12$\pm$1.34                   & \textbf{4.17$\pm$1.18}             & 5.74$\pm$1.47                    & 9.63$\pm$2.03                    \\
                          & Node2vec               & \textbf{93.83$\pm$2.15}           & 89.69$\pm$2.30                   & \multicolumn{1}{l|}{86.87$\pm$3.21} & 71.43$\pm$2.60                   & 44.44$\pm$2.08                     & 10.86$\pm$2.26                    & 6.08$\pm$1.49                     & 4.21$\pm$1.20                      & \textbf{3.36$\pm$0.92}           & 4.59$\pm$1.41                    
                          \\ \hline
\multirow{3}{*}{Yeast}    & GAE                    & 68.69$\pm$1.85                    & 60.64$\pm$2.01                    & \multicolumn{1}{l|}{54.84$\pm$3.11} & \textbf{69.52$\pm$2.17}          & 5.03$\pm$4.36                     & 73.13$\pm$8.44                    & 48.24$\pm$6.12                    & \textbf{41.82$\pm$5.46}            & 46.78$\pm$3.81                   & 67.27$\pm$7.62                    \\
                          & Deepwalk               & \textbf{94.57$\pm$2.42}           & 93.42$\pm$2.01                    & \multicolumn{1}{l|}{90.82$\pm$2.67} & 94.21$\pm$2.04                  & 76.67$\pm$3.81                    & 45.10$\pm$5.79                   & 31.71$\pm$5.01                    & 26.14$\pm$3.81                    & \textbf{22.77$\pm$2.40}          & 43.60$\pm$4.49                   \\
                          & Node2vec               & \textbf{96.76$\pm$1.24}           & 94.90$\pm$1.41                    & \multicolumn{1}{l|}{89.79$\pm$3.21} & 95.51$\pm$1.31                  & 76.67$\pm$3.55                     & 48.35$\pm$3.24                   & 33.57$\pm$2.80                   & 26.81$\pm$3.56                    & \textbf{22.46$\pm$2.87}          & 45.20$\pm$3.99
                          \\ \hline
\multirow{3}{*}{Facebook} & GAE                    & \textbf{56.99$\pm$3.54}           & 48.96$\pm$3.71                    & \multicolumn{1}{l|}{42.71$\pm$4.06} & 55.84$\pm$3.22                   & 1.33$\pm$1.14                      & 210.61$\pm$13.47                  & 168.22$\pm$10.57                   & \textbf{120.57$\pm$9.54}           & 134.97$\pm$9.86                  & 189.94$\pm$10.24                   \\
                          & Deepwalk               & 97.54$\pm$2.46          & 94.85$\pm$1.88                    & \multicolumn{1}{l|}{90.72$\pm$2.34} & \textbf{97.66$\pm$2.34}                     & 95.13$\pm$3.74                       & 93.42$\pm$6.47                   & 75.10$\pm$5.21                   & \textbf{68.66$\pm$5.11}            & 99.26$\pm$7.15                  & 93.33$\pm$6.57                   \\
                          & Node2vec               & \textbf{95.86$\pm$3.17}           & 92.90$\pm$3.55                   & \multicolumn{1}{l|}{88.47$\pm$5.66} & 94.14$\pm$3.62                    & 89.47$\pm$3.80                       & 93.54$\pm$8.14                   & 74.80$\pm$6.24                    & \textbf{67.98$\pm$5.93}            & 80.10$\pm$6.32                   & 94.33$\pm$7.84
                           \\ \hline

\end{tabular}}
\label{tab:5}%
\end{table*}

\subsubsection{Link prediction attack} Here we introduce the performance and transferability of GraphAttacker in link prediction attack.

\setlength{\parskip}{0\baselineskip}
\noindent\textbf{Model configurations.} {In the link prediction task, we keep the same dataset division ratio as in Section \ref{graphatt}.
We use GAE\cite{kipf2016variational} as the target attack model. GAE consists of a GCN encoder and a simple \emph{inner product} decoder. The structure and parameter settings of the encoder are the same as the feature extractor introduced in Section \ref{sec4.4.1}. The encoder learns the graph embedding $Z$. Then, the decoder calculates the inner product of $Z$ to obtain the probability of the existence of links.}

\setlength{\parskip}{0\baselineskip}
\noindent\textbf{Attack strategy.} From {TABLE~\ref{tab:1}}, we can see that the datasets used for link prediction attack are similar to those in node classification and only contain one large-size graph. Therefore, we select $K$-hop subgraphs of nodes $v_i$ and $v_j$, from the target link as our attack targets. The datasets of Yeast and Facebook have large node degrees, $G_{sub}$ already has more than thousands of nodes when $K=2$. {We choose $K=3$ for NS and Cora, and $K=1$ for Yeast and Facebook to reduce the attack cost. Since there are no node attributes for most of the datasets used for link prediction attacks, we only conduct graph structure attacks for the convenience of comparison.}

\setlength{\parskip}{0\baselineskip}
{We find that unlimited attack achieves a higher ASR than direct attack in most cases in link prediction attack, which is different from the situation in node classification attack. We speculate that it may be challenging to cover up the influence of the other link to the target link in a link prediction attack by only directly adding links to the nodes at both ends of the target link. Simultaneously, the unlimited attack affects the target link itself and affects the features of the original links around it}, making it have a better attack effect. In this case, the impact of the subtle difference between subgraph attack and whole graph attack can be ignored.

\noindent\textbf{Attack performance.}
In this part, we compare unlimited attack with other attack methods in link prediction on {TABLE~\ref{tab:5}}, including B-GA, D-GA, and S-GA. In most cases, B-GA can achieve the highest ASR.
However, under the same constraints, the AML of IGA is lower than B-GA. {Considering that IGA performs an adversarial attack by modifying the link with the largest gradient}, we think this result is reasonable. Similar to the result in {section \ref{section:5.5.1}}, we can also observe that S-GA and D-GA have smaller AML when attack link prediction task, although their ASR cannot reach the highest. In our experiments, the ASR of B-GA's transfer attack on the traditional graph embedding methods is higher than that of other baselines, which shows that GraphAttacker has stronger transferability.
\setlength{\parskip}{-0.5\baselineskip}

{\subsection{Attack Performance Under Possible Defenses}
\setlength{\parskip}{0\baselineskip}
\noindent To further verify the advantages of GraphAttacker, we take the node classification attack as an example to investigate the attack performance of different attack methods under several possible defenses. Here, we conduct attack experiments under three different types of defense methods, including the classic adversarial training, the model structure-based GNNGuard\cite{zhang2020gnnguard}, and the model parameter-based robust training\cite{bojchevski2019certifiable}. Specifically, in the adversarial training for different node classification attack methods, we first generate ten adversarial training examples for each target node and then mix them with the original one to retrain the GCN. Due to the multiple attack strategies in GraphAttacker, we can combine multiple modification strategies and attack scale to generate diversiform adversarial examples for adversarial training, termed GA. After the GCN is retrained, we attack the retrained GCN by other newly generated adversarial examples by the same attack. For GNNGuard and robust training, we conduct experiments based on the source code released by the authors and their suggested parameter settings.

\begin{table*}[]
\centering
\caption{The ASR (mean and 95\% confidence interval) of different attack methods under various defense methods. GA is the GraphAttacker considering various attack strategies, S-GA denotes the GraphAttacker for direct attack under SMR constraints. We use \textbf{bold} to highlight wins. }
\resizebox{178mm}{30mm}{
\begin{tabular}{cccccccccc}
\hline
\multirow{2}{*}{Dataset}   & \multirow{2}{*}{Defense} & \multicolumn{8}{c}{ASR(\%)}                                                                             \\ \cline{3-10}
                           &                          & GA             & S-GA           & DICE  & NETTACK & GF-Attack & GUA   & IG-JSMA        & SGA            \\ \hline
\multirow{4}{*}{Pol.Blogs} & No   defense             & \textbf{89.99$\pm$1.40} & 51.28$\pm$2.85    & 50.27$\pm$4.43 & 82.97$\pm$1.74   & 19.89$\pm$3.41   & 46.42$\pm$3.41 & 89.14$\pm$1.74    & 86.68$\pm$1.34  \\
                           & Adversarial   training   & \textbf{59.47$\pm$2.54} & 37.61$\pm$2.19    & 33.34$\pm$4.47 & 27.54$\pm$2.26   & 15.54$\pm$1.70   & 5.14$\pm$1.93  & 26.45$\pm$2.55    & 38.74$\pm$2.23  \\
                           & GNNGuard                 & 42.67$\pm$3.41    & \textbf{44.43$\pm$1.72} & 11.82$\pm$3.56 & 19.60$\pm$1.82    & 14.73$\pm$2.31     & 16.43$\pm$1.41 & 12.47$\pm$0.86   & 39.45$\pm$2.04          \\
                           & Robust   training        & 56.23$\pm$0.84    & 30.67$\pm$0.69   & 24.82$\pm$1.14 & 53.06$\pm$0.81  & 15.93$\pm$1.92     & 24.97$\pm$1.28 & \textbf{60.43$\pm$0.46} & 55.67$\pm$0.86    \\ \hline
\multirow{4}{*}{Cora}      & No   defense             & \textbf{98.41$\pm$1.59} & 78.26$\pm$2.23    & 52.36$\pm$2.49 & 91.04$\pm$1.83   & 80.67$\pm$1.88     &  84.68$\pm$3.08  &97.64$\pm$1.56  &94.74$\pm$2.14        \\
                           & Adversarial   training   & \textbf{66.43$\pm$2.68} & 53.74$\pm$1.85    & 34.24$\pm$2.27 & 36.51$\pm$2.02   & 29.29$\pm$1.84     & 12.41$\pm$1.66 & 32.67$\pm$1.37  & 46.15$\pm$2.36          \\
                           & GNNGuard                 & 54.07$\pm$3.18    & \textbf{68.46$\pm$2.24} & 14.22$\pm$3.37 & 29.43$\pm$1.82   & 26.74$\pm$1.47     & 28.43$\pm$1.76 & 24.54$\pm$1.08   & 43.96$\pm$1.95          \\
                           & Robust   training        & 66.43$\pm$0.76          & 41.50$\pm$0.71          & 26.61$\pm$1.32 & 61.04$\pm$0.86   & 31.49$\pm$1.52     & 38.67$\pm$1.17 & \textbf{69.42$\pm$0.38} & 62.67$\pm$0.84          \\ \hline
\multirow{4}{*}{Citeseer}  & No   defense             & \textbf{98.53$\pm$1.47} & 81.12$\pm$2.34  &67.14$\pm$3.59  &86.01$\pm$1.65  &58.47$\pm$2.89  &78.16$\pm$4.13  &97.16$\pm$1.57  &92.79$\pm$2.26        \\
                           & Adversarial   tTraining   & \textbf{56.57$\pm$2.14} & 42.84$\pm$1.49          & 37.52$\pm$2.53 & 38.41$\pm$2.60   & 32.41$\pm$1.98     & 11.86$\pm$3.17 & 31.72$\pm$1.43  & 45.84$\pm$2.54          \\
                           & GNNGuard                 & 43.75$\pm$2.81          & \textbf{63.72$\pm$1.63} & 16.98$\pm$3.17 & 31.87$\pm$2.34   & 27.72$\pm$2.35     & 32.36$\pm$2.96 & 28.74$\pm$1.47   & 41.77$\pm$2.62          \\
                           & Robust   training        & 49.10$\pm$0.74          & 33.47$\pm$0.66          & 27.59$\pm$1.46 & 47.68$\pm$0.76   & 32.16$\pm$1.58    & 36.81$\pm$1.37 & \textbf{61.52$\pm$0.51} & 47.71$\pm$1.31          \\ \hline
\multirow{4}{*}{Pubmed}    & No   defense             & \textbf{98.62$\pm$1.27} & 66.41$\pm$2.16   &51.44$\pm$3.63  &82.14$\pm$1.54  &52.47$\pm$2.34  &85.45$\pm$3.47  &98.33$\pm$1.41  &95.43$\pm$2.04   \\
                           & Adversarial   training   & \textbf{64.57$\pm$2.84} & 37.40$\pm$2.41          & 35.14$\pm$3.44 & 34.54$\pm$1.63   & 26.54$\pm$1.84     & 13.05$\pm$3.06 & 34.08$\pm$1.35          & 46.92$\pm$2.47          \\
                           & GNNGuard                 & 48.43$\pm$2.47          & \textbf{57.60$\pm$1.88} & 19.12$\pm$3.39 & 29.63$\pm$2.03   & 26.06$\pm$1.76     & 30.93$\pm$2.78 & 31.55$\pm$1.54          & 44.20$\pm$2.26         \\
                           & Robust   training        & 57.25$\pm$0.63          & 31.58$\pm$0.69          & 27.41$\pm$2.10 & 51.16$\pm$0.82   & 32.41$\pm$1.62     & 37.51$\pm$2.18 & 57.06$\pm$0.49          & \textbf{58.34$\pm$1.06} \\ \hline
\end{tabular}\label{tab:7}}
\end{table*}

TABLE~\ref{tab:7} shows the attack performance of different attack methods on GCN with or without possible defenses. Compared with other methods, GA can still achieve satisfying attack effects when undergoing adversarial training. We can observe that adversarial training reduces the ASR of NETTACK, GF attack, IG-JSMA, and SGA by more than 50\%. It proves that adversarial training has a strong defense ability in attack methods obtained perturbation candidate sets through specific attack strategies such as greedy methods and gradient learning. In addition, since GUA aims to achieve a general attack on different nodes through the same group of anchor nodes, it is an extreme example of the lack of diversity of adversarial examples. The adversarial training significantly reduces the ASR of GUA, which indicates that GUA is limited by the lack of example diversity and is easier to be defended by adversarial training. Although DICE has limited attack effects, it can generate diversiform adversarial examples due to the random attack strategy. This can help DICE reduce the impact of adversarial training. In the attack process of GraphAttacker, multiple attack strategies and the strategy of randomly adding nodes to the K-hop subgraph ensure the diversity and effectiveness of the generated adversarial examples. The attack performance of GraphAttacker and other attack methods under adversarial training further proves the importance of diversiform generation of adversarial examples.

Under the defense of GNNGuard, S-GA shows better and more stable attack performance due to the limitation of SMR. GNNGuard learns how to assign higher weights to edges connecting similar nodes, and implements defense by pruning edges between unrelated nodes. Compared with other attack methods, only S-GA restricts the change of node similarity, which makes the adversarial perturbations generated by S-GA more difficult to be detected by GNNGuard. Robust training increases the number of certifiably robust nodes by optimizing the loss function of the GNN model. In this case, IG-JSMA, which generates the largest impact perturbations through the integral gradient of the target model, seems to perform better, while GA still exhibits competitive attack performance.

All in all, diversified adversarial example generation strategies and constraints on node similarity bring better attack performance to GraphAttacker even when faced with different defense methods. Diversified adversarial examples can effectively reduce the defense ability of adversarial training, while the node similarity constraint can help GraphAttacker avoid the detection of node similarity-based defense methods.

\setlength{\parskip}{1\baselineskip}

\subsection{Summary and Insight}
\setlength{\parskip}{-0.3\baselineskip}
{ Finally, we summarize the experiment results of varying graph attack tasks.} We put forward our insights for the impact of different graph datasets on graph analysis task attacks (question \textbf{RQ3}) and the attack performance of different graph analysis tasks (question \textbf{RQ4}).
\setlength{\parskip}{0\baselineskip}

For question \textbf{RQ3}, since the datasets we used in the link prediction attack only contain graph structure information, we compare the attack performance of different modification strategies in node classification attack and graph classification attack. In the datasets
of Cora, Citeseer, Pol.Blog, and Pubmed most node attributes are 0, and their graph structure is relatively dense, containing more information. Therefore, in node classification attack, graph structure attack has a better attack effect than node attribute attack.
{In the datasets of PROTEINS, DD, NCI, and IMDB-BINARY the graph structure of each dataset is also dense, and the graph structure attack also achieves a better attack effect.} In PROTEINS-full and ENZYMES,
attacks on continuous node attributes have achieved a more significant attack effect.
{We conclude that graph analysis models may pay more attention to the graph feature (graph structure/node attributes), containing more information.} Relatively dense information may be the weak point of the datasets. Similarly, {Wang et al.}\cite{wang2020gcn} extracted the embedding of node attributes by constructing an attribute graph. They proved that the graph structure and node attributes in different graph datasets have different importance in the GNN model, supporting our insights.

For question \textbf{RQ4}, under the same constraints, graph classification attack is more difficult to achieve successful attack than node classification attack.
{Considering the scope of information, the node classification model often only needs to consider the neighborhood information of the target node. The graph classification model aggregates the information of all nodes in the graph, so it usually requires a larger attack budget.}
Like the node classification task, the link prediction task also predicts an instance (link) in the graph. However, it is more difficult to attack. The link prediction model calculates the probability of a link between two nodes.
{A successful attack usually requires a significant difference in the predicted probability. We believe that the classification boundary of the link prediction task is more difficult to cross than node classification.}

\section{CONCLUSION}\label{sec6}
We have proposed the first general attack framework for multiple graph analysis tasks to generate diversiform adversarial examples and better understand the attack characteristics of different graph analysis tasks. Based on the idea of GAN, we use a three-player game of MAG, SD, and AD to achieve general attacks under multiple modification strategies, attack scale, and stealthiness constraints. We have explored the possible shortcoming of the existing perturbation constraints and preserved the average similarity distribution of nodes in the adversarial examples, which have further improved the stealthiness of perturbations.  Experiments on various benchmark datasets demonstrate that GraphAttacker can achieve state-of-the-art attack performance on graph analysis tasks of node classification, graph classification, and link prediction, no matter the adversarial training is conducted or not. Moreover, when conducting general attacks, we have analyzed the unique characteristics of each task and put forward our insights.

The robustness of the graph analysis task model is an important issue. This work has analyzed the different weaknesses for different graph analysis tasks. For future work, we hope to apply the GAN to the defense of graph analysis tasks, and achieve a more targeted defense based on the unique characteristics of different graph analysis tasks.


\section*{Acknowledgments}

The authors would like to thank the National Natural Science Foundation of China under Grant No. 62072406, the Natural Science Foundation of Zhejiang Province under Grant No. LY19F020025, the Key Laboratory of the Public Security Ministry Open Project in 2020 under Grant No. 2020DSJSYS001, the Key R\&D Projects in Zhejiang Province under Grant No.2021C01117, the 2020 Industrial Internet Innovation Development Project under Grant No.TC200H01V, the Ten Thousand Talents Program" Science and Technology Innovation Leading Talent Project in Zhejiang Province under Grant No.2020R52011, the National Key Research and Development Program of China under Grant No. 2018AAA0100801.


%

%
%
%
%
%

\ifCLASSOPTIONcaptionsoff
  \newpage
\fi

\bibliographystyle{IEEEtran}
\bibliography{refer}

\begin{thebibliography}{10}
\providecommand{\url}[1]{#1}
\csname url@samestyle\endcsname
\providecommand{\newblock}{\relax}
\providecommand{\bibinfo}[2]{#2}
\providecommand{\BIBentrySTDinterwordspacing}{\spaceskip=0pt\relax}
\providecommand{\BIBentryALTinterwordstretchfactor}{4}
\providecommand{\BIBentryALTinterwordspacing}{\spaceskip=\fontdimen2\font plus
\BIBentryALTinterwordstretchfactor\fontdimen3\font minus
  \fontdimen4\font\relax}
\providecommand{\BIBforeignlanguage}[2]{{%
\expandafter\ifx\csname l@#1\endcsname\relax
\typeout{** WARNING: IEEEtran.bst: No hyphenation pattern has been}%
\typeout{** loaded for the language `#1'. Using the pattern for}%
\typeout{** the default language instead.}%
\else
\language=\csname l@#1\endcsname
\fi
#2}}
\providecommand{\BIBdecl}{\relax}
\BIBdecl

\bibitem{8411156}
D.~{Yang}, M.~{Liu}, Y.~{Zhang}, D.~{Lin}, Z.~{Fan}, and G.~{Chen}, ``Henneberg
  growth of social networks: Modeling the facebook,'' \emph{IEEE Transactions
  on Network Science and Engineering}, vol.~7, no.~2, pp. 701--712, 2020.

\bibitem{wang2018billion}
J.~Wang, P.~Huang, H.~Zhao, Z.~Zhang, B.~Zhao, and D.~L. Lee, ``Billion-scale
  commodity embedding for e-commerce recommendation in alibaba,'' in
  \emph{Proceedings of the 24th ACM SIGKDD International Conference on
  Knowledge Discovery \& Data Mining}, 2018, pp. 839--848.

\bibitem{8472785}
G.~{Mangioni}, G.~{Jurman}, and M.~{De Domenico}, ``Multilayer flows in
  molecular networks identify biological modules in the human proteome,''
  \emph{IEEE Transactions on Network Science and Engineering}, vol.~7, no.~1,
  pp. 411--420, 2020.

\bibitem{latora2002boston}
V.~Latora and M.~Marchiori, ``Is the boston subway a small-world network?''
  \emph{Physica A: Statistical Mechanics and its Applications}, vol. 314, no.
  1-4, pp. 109--113, 2002.

\bibitem{hamilton2017inductive}
W.~L. Hamilton, R.~Ying, and J.~Leskovec, ``Inductive representation learning
  on large graphs,'' \emph{arXiv preprint arXiv:1706.02216}, 2017.

\bibitem{kipf2017semi}
\BIBentryALTinterwordspacing
T.~N. Kipf and M.~Welling, ``Semi-supervised classification with graph
  convolutional networks,'' in \emph{International Conference on Learning
  Representations}, ser. ICLR '17, Toulon, France, 2017. [Online]. Available:
  \url{https://openreview.net/forum?id=SJU4ayYgl&noteId=SJU4ayYgl}
\BIBentrySTDinterwordspacing

\bibitem{Hu*2020Strategies}
\BIBentryALTinterwordspacing
W.~Hu, B.~Liu, J.~Gomes, M.~Zitnik, P.~Liang, V.~Pande, and J.~Leskovec,
  ``Strategies for pre-training graph neural networks,'' in \emph{International
  Conference on Learning Representations}, ser. ICLR'20, 2020. [Online].
  Available: \url{https://openreview.net/forum?id=HJlWWJSFDH}
\BIBentrySTDinterwordspacing

\bibitem{perozzi2014deepwalk}
B.~Perozzi, R.~Al-Rfou, and S.~Skiena, ``Deepwalk: Online learning of social
  representations,'' in \emph{Proceedings of the 20th ACM SIGKDD international
  conference on Knowledge discovery and data mining}, 2014, pp. 701--710.

\bibitem{grover2016node2vec}
A.~Grover and J.~Leskovec, ``node2vec: Scalable feature learning for
  networks,'' in \emph{Proceedings of the 22nd ACM SIGKDD international
  conference on Knowledge discovery and data mining}, 2016, pp. 855--864.

\bibitem{tang2015pte}
J.~Tang, M.~Qu, and Q.~Mei, ``Pte: Predictive text embedding through
  large-scale heterogeneous text networks,'' in \emph{Proceedings of the 21th
  ACM SIGKDD international conference on knowledge discovery and data mining},
  2015, pp. 1165--1174.

\bibitem{wang2016linked}
S.~Wang, J.~Tang, C.~Aggarwal, and H.~Liu, ``Linked document embedding for
  classification,'' in \emph{Proceedings of the 25th ACM international on
  conference on information and knowledge management}, 2016, pp. 115--124.

\bibitem{ying2018hierarchical}
R.~Ying, J.~You, C.~Morris, X.~Ren, W.~L. Hamilton, and J.~Leskovec,
  ``Hierarchical graph representation learning with differentiable pooling,''
  \emph{arXiv preprint arXiv:1806.08804}, 2018.

\bibitem{lee2019self}
J.~Lee, I.~Lee, and J.~Kang, ``Self-attention graph pooling,''
  \emph{International Conference on Machine Learning}, pp. 3734--¨C3743, 2019.

\bibitem{wang2017signed}
S.~Wang, J.~Tang, C.~Aggarwal, Y.~Chang, and H.~Liu, ``Signed network embedding
  in social media,'' in \emph{Proceedings of the 2017 SIAM international
  conference on data mining}.\hskip 1em plus 0.5em minus 0.4em\relax SIAM,
  2017, pp. 327--335.

\bibitem{tian2014learning}
F.~Tian, B.~Gao, Q.~Cui, E.~Chen, and T.-Y. Liu, ``Learning deep
  representations for graph clustering,'' in \emph{Proceedings of the AAAI
  Conference on Artificial Intelligence}, vol.~28, no.~1, 2014.

\bibitem{7560670}
K.~{Allab}, L.~{Labiod}, and M.~{Nadif}, ``A semi-nmf-pca unified framework for
  data clustering,'' \emph{IEEE Transactions on Knowledge and Data
  Engineering}, vol.~29, no.~1, pp. 2--16, 2017.

\bibitem{7769223}
V.~{Lyzinski}, M.~{Tang}, A.~{Athreya}, Y.~{Park}, and C.~E. {Priebe},
  ``Community detection and classification in hierarchical stochastic
  blockmodels,'' \emph{IEEE Transactions on Network Science and Engineering},
  vol.~4, no.~1, pp. 13--26, 2017.

\bibitem{pareja2020evolvegcn}
A.~Pareja, G.~Domeniconi, J.~Chen, T.~Ma, T.~Suzumura, H.~Kanezashi, T.~Kaler,
  T.~Schardl, and C.~Leiserson, ``Evolvegcn: Evolving graph convolutional
  networks for dynamic graphs,'' in \emph{Proceedings of the AAAI Conference on
  Artificial Intelligence}, vol.~34, no.~04, 2020, pp. 5363--5370.

\bibitem{wang2019heterogeneous}
S.~Wang, Z.~Chen, X.~Yu, D.~Li, J.~Ni, L.-A. Tang, J.~Gui, Z.~Li, H.~Chen, and
  P.~S. Yu, ``Heterogeneous graph matching networks,'' \emph{arXiv preprint
  arXiv:1910.08074}, 2019.

\bibitem{8923033}
S.~A. {Al-Zboon}, S.~K. {Tawalbeh}, H.~{Ai-Jarrah}, M.~{Al-Asa'd}, M.~{Hammad},
  and M.~{Al-Smadi}, ``Resolving conflict of interests in recommending
  reviewers for academic publications using link prediction techniques,'' in
  \emph{2019 2nd International Conference on new Trends in Computing Sciences
  (ICTCS)}, 2019, pp. 1--6.

\bibitem{8424608}
M.~{Lu}, Z.~{Qu}, M.~{Wang}, and Z.~{Qin}, ``Recommending authors and papers
  based on acttm community and bilayer citation network,'' \emph{China
  Communications}, vol.~15, no.~7, pp. 111--130, 2018.

\bibitem{krebs2002mapping}
V.~E. Krebs, ``Mapping networks of terrorist cells,'' \emph{Connections},
  vol.~24, no.~3, pp. 43--52, 2002.

\bibitem{7938634}
S.~{Ji}, T.~{Wang}, J.~{Chen}, W.~{Li}, P.~{Mittal}, and R.~{Beyah}, ``De-sag:
  On the de-anonymization of structure-attribute graph data,'' \emph{IEEE
  Transactions on Dependable and Secure Computing}, vol.~16, no.~4, pp.
  594--607, 2019.

\bibitem{9101713}
S.~{Ji}, Q.~{Gu}, H.~{Weng}, Q.~{Liu}, P.~{Zhou}, J.~{Chen}, Z.~{Li},
  R.~{Beyah}, and T.~{Wang}, ``De-health: All your online health information
  are belong to us,'' in \emph{2020 IEEE 36th International Conference on Data
  Engineering (ICDE)}, 2020, pp. 1609--1620.

\bibitem{zugner2018adversarial}
D.~Z{\"u}gner, A.~Akbarnejad, and S.~G{\"u}nnemann, ``Adversarial attacks on
  neural networks for graph data,'' in \emph{Proceedings of the 24th ACM SIGKDD
  International Conference on Knowledge Discovery \& Data Mining}, 2018, pp.
  2847--2856.

\bibitem{chen2018fast}
J.~Chen, Y.~Wu, X.~Xu, Y.~Chen, H.~Zheng, and Q.~Xuan, ``Fast gradient attack
  on network embedding,'' \emph{arXiv preprint arXiv:1809.02797}, 2018.

\bibitem{chang2020restricted}
H.~Chang, Y.~Rong, T.~Xu, W.~Huang, H.~Zhang, P.~Cui, W.~Zhu, and J.~Huang, ``A
  restricted black-box adversarial framework towards attacking graph embedding
  models,'' in \emph{Proceedings of the AAAI Conference on Artificial
  Intelligence}, vol.~34, no.~04, 2020, pp. 3389--3396.

\bibitem{xi2020graph}
Z.~Xi, R.~Pang, S.~Ji, and T.~Wang, ``Graph backdoor,'' \emph{USENIX Security},
  2021.

\bibitem{zhang2020backdoor}
Z.~Zhang, J.~Jia, B.~Wang, and N.~Z. Gong, ``Backdoor attacks to graph neural
  networks,'' \emph{arXiv preprint arXiv:2006.11165}, 2020.

\bibitem{8714065}
J.~{Chen}, L.~{Chen}, Y.~{Chen}, M.~{Zhao}, S.~{Yu}, Q.~{Xuan}, and X.~{Yang},
  ``Ga-based q-attack on community detection,'' \emph{IEEE Transactions on
  Computational Social Systems}, vol.~6, no.~3, pp. 491--503, 2019.

\bibitem{ma2020towards}
J.~Ma, S.~Ding, and Q.~Mei, ``Towards more practical adversarial attacks on
  graph neural networks,'' \emph{Advances in neural information processing
  systems}, 2020.

\bibitem{dai2019adversarial}
Q.~Dai, X.~Shen, L.~Zhang, Q.~Li, and D.~Wang, ``Adversarial training methods
  for network embedding,'' in \emph{The World Wide Web Conference}, 2019, pp.
  329--339.

\bibitem{8924766}
F.~{Feng}, X.~{He}, J.~{Tang}, and T.~{Chua}, ``Graph adversarial training:
  Dynamically regularizing based on graph structure,'' \emph{IEEE Transactions
  on Knowledge and Data Engineering}, pp. 1--1, 2019.

\bibitem{zhang2019comparing}
Y.~Zhang, S.~Khan, and M.~Coates, ``Comparing and detecting adversarial attacks
  for graph deep learning,'' in \emph{Proc. Representation Learning on Graphs
  and Manifolds Workshop, Int. Conf. Learning Representations, New Orleans, LA,
  USA}, 2019.

\bibitem{9305289}
J.~{Chen}, X.~{Lin}, H.~{Xiong}, Y.~{Wu}, H.~{Zheng}, and Q.~{Xuan},
  ``Smoothing adversarial training for gnn,'' \emph{IEEE Transactions on
  Computational Social Systems}, pp. 1--12, 2020.

\bibitem{entezari2020all}
N.~Entezari, S.~A. Al-Sayouri, A.~Darvishzadeh, and E.~E. Papalexakis, ``All
  you need is low (rank) defending against adversarial attacks on graphs,'' in
  \emph{Proceedings of the 13th International Conference on Web Search and Data
  Mining}, 2020, pp. 169--177.

\bibitem{tang2020transferring}
X.~Tang, Y.~Li, Y.~Sun, H.~Yao, P.~Mitra, and S.~Wang, ``Transferring
  robustness for graph neural network against poisoning attacks,'' in
  \emph{Proceedings of the 13th International Conference on Web Search and Data
  Mining}, 2020, pp. 600--608.

\bibitem{9006004}
T.~{Takahashi}, ``Indirect adversarial attacks via poisoning neighbors for
  graph convolutional networks,'' in \emph{2019 IEEE International Conference
  on Big Data (Big Data)}, 2019, pp. 1395--1400.

\bibitem{ioannidis2019graphsac}
V.~N. Ioannidis, D.~Berberidis, and G.~B. Giannakis, ``Graphsac: Detecting
  anomalies in large-scale graphs,'' \emph{arXiv preprint arXiv:1910.09589},
  2019.

\bibitem{zugner2019adversarial}
D.~Z{\"u}gner and S.~G{\"u}nnemann, ``Adversarial attacks on graph neural
  networks via meta learning,'' \emph{arXiv preprint arXiv:1902.08412}, 2019.

\bibitem{wu2019adversarial}
H.~Wu, C.~Wang, Y.~Tyshetskiy, A.~Docherty, K.~Lu, and L.~Zhu, ``Adversarial
  examples on graph data: Deep insights into attack and defense,'' \emph{arXiv
  preprint arXiv:1903.01610}, 2019.

\bibitem{zhang2020gnnguard}
X.~Zhang and M.~Zitnik, ``Gnnguard: Defending graph neural networks against
  adversarial attacks,'' \emph{arXiv preprint arXiv:2006.08149}, 2020.

\bibitem{jin2020graph}
W.~Jin, Y.~Ma, X.~Liu, X.~Tang, S.~Wang, and J.~Tang, ``Graph structure
  learning for robust graph neural networks,'' in \emph{Proceedings of the 26th
  ACM SIGKDD International Conference on Knowledge Discovery \& Data Mining},
  2020, pp. 66--74.

\bibitem{wu2020phishers}
J.~Wu, Q.~Yuan, D.~Lin, W.~You, W.~Chen, C.~Chen, and Z.~Zheng, ``Who are the
  phishers? phishing scam detection on ethereum via network embedding,''
  \emph{IEEE Transactions on Systems, Man, and Cybernetics: Systems}, 2020.

\bibitem{zhang2021blockchain}
D.~Zhang and J.~Chen, ``Blockchain phishing scam detection via multi-channel
  graph classification,'' \emph{arXiv preprint arXiv:2108.08456}, 2021.

\bibitem{lin2020t}
D.~Lin, J.~Wu, Q.~Yuan, and Z.~Zheng, ``T-edge: Temporal weighted multidigraph
  embedding for ethereum transaction network analysis,'' \emph{Frontiers in
  Physics}, vol.~8, p. 204, 2020.

\bibitem{wang2018attack}
X.~Wang, M.~Cheng, J.~Eaton, C.-J. Hsieh, and F.~Wu, ``Attack graph
  convolutional networks by adding fake nodes,'' \emph{arXiv preprint
  arXiv:1810.10751}, 2018.

\bibitem{zang2020graph}
X.~Zang, Y.~Xie, J.~Chen, and B.~Yuan, ``Graph universal adversarial attacks: A
  few bad actors ruin graph learning models,'' \emph{arXiv preprint
  arXiv:2002.04784}, 2020.

\bibitem{li2021adversarial}
J.~Li, T.~Xie, C.~Liang, F.~Xie, X.~He, and Z.~Zheng, ``Adversarial attack on
  large scale graph,'' \emph{IEEE Transactions on Knowledge and Data
  Engineering}, 2021.

\bibitem{dai2018adversarial}
H.~Dai, H.~Li, T.~Tian, X.~Huang, L.~Wang, J.~Zhu, and L.~Song, ``Adversarial
  attack on graph structured data,'' in \emph{International Conference on
  Machine Learning}.\hskip 1em plus 0.5em minus 0.4em\relax PMLR, 2018, pp.
  1115--1124.

\bibitem{tang2020adversarial}
H.~Tang, G.~Ma, Y.~Chen, L.~Guo, W.~Wang, B.~Zeng, and L.~Zhan, ``Adversarial
  attack on hierarchical graph pooling neural networks,'' \emph{arXiv preprint
  arXiv:2005.11560}, 2020.

\bibitem{9141291}
J.~{Chen}, X.~{Lin}, Z.~{Shi}, and Y.~{Liu}, ``Link prediction adversarial
  attack via iterative gradient attack,'' \emph{IEEE Transactions on
  Computational Social Systems}, vol.~7, no.~4, pp. 1081--1094, 2020.

\bibitem{kipf2016variational}
T.~N. Kipf and M.~Welling, ``Variational graph auto-encoders,'' \emph{arXiv
  preprint arXiv:1611.07308}, 2016.

\bibitem{chen2019time}
J.~Chen, J.~Zhang, Z.~Chen, M.~Du, and Q.~Xuan, ``Time-aware gradient attack on
  dynamic network link prediction,'' \emph{arXiv preprint arXiv:1911.10561},
  2019.

\bibitem{9172881}
S.~{Yu}, J.~{Zheng}, J.~{Chen}, Q.~{Xuan}, and Q.~{Zhang}, ``Unsupervised
  euclidean distance attack on network embedding,'' in \emph{2020 IEEE Fifth
  International Conference on Data Science in Cyberspace (DSC)}, 2020, pp.
  71--77.

\bibitem{sun2019virtual}
K.~Sun, Z.~Lin, H.~Guo, and Z.~Zhu, ``Virtual adversarial training on graph
  convolutional networks in node classification,'' in \emph{Chinese Conference
  on Pattern Recognition and Computer Vision (PRCV)}.\hskip 1em plus 0.5em
  minus 0.4em\relax Springer, 2019, pp. 431--443.

\bibitem{zhu2019robust}
D.~Zhu, Z.~Zhang, P.~Cui, and W.~Zhu, ``Robust graph convolutional networks
  against adversarial attacks,'' in \emph{Proceedings of the 25th ACM SIGKDD
  International Conference on Knowledge Discovery \& Data Mining}, 2019, pp.
  1399--1407.

\bibitem{bojchevski2019certifiable}
A.~Bojchevski and S.~G{\"u}nnemann, ``Certifiable robustness to graph
  perturbations,'' \emph{arXiv preprint arXiv:1910.14356}, 2019.

\bibitem{DBLP:conf/aaai/LiuFDQC19}
\BIBentryALTinterwordspacing
P.~Liu, J.~Fu, Y.~Dong, X.~Qiu, and J.~C.~K. Cheung, ``Learning multi-task
  communication with message passing for sequence learning,'' in \emph{The
  Thirty-Third {AAAI} Conference on Artificial Intelligence, {AAAI} 2019, The
  Thirty-First Innovative Applications of Artificial Intelligence Conference,
  {IAAI} 2019, The Ninth {AAAI} Symposium on Educational Advances in Artificial
  Intelligence, {EAAI} 2019, Honolulu, Hawaii, USA, January 27 - February 1,
  2019}.\hskip 1em plus 0.5em minus 0.4em\relax {AAAI} Press, 2019, pp.
  4360--4367. [Online]. Available:
  \url{https://doi.org/10.1609/aaai.v33i01.33014360}
\BIBentrySTDinterwordspacing

\bibitem{DBLP:conf/iclr/XuHLJ19}
\BIBentryALTinterwordspacing
K.~Xu, W.~Hu, J.~Leskovec, and S.~Jegelka, ``How powerful are graph neural
  networks?'' in \emph{7th International Conference on Learning
  Representations, {ICLR} 2019, New Orleans, LA, USA, May 6-9, 2019}.\hskip 1em
  plus 0.5em minus 0.4em\relax OpenReview.net, 2019. [Online]. Available:
  \url{https://openreview.net/forum?id=ryGs6iA5Km}
\BIBentrySTDinterwordspacing

\bibitem{holtz2019multi}
C.~Holtz, O.~Atan, R.~Carey, and T.~Jain, ``Multi-task learning on graphs with
  node and graph level labels,'' \emph{SAGE}, vol.~4, no. 76.84, pp. 62--4,
  2019.

\bibitem{DBLP:conf/nips/ZhangWY18}
\BIBentryALTinterwordspacing
Y.~Zhang, Y.~Wei, and Q.~Yang, ``Learning to multitask,'' in \emph{Advances in
  Neural Information Processing Systems 31: Annual Conference on Neural
  Information Processing Systems 2018, NeurIPS 2018, December 3-8, 2018,
  Montr{\'{e}}al, Canada}, S.~Bengio, H.~M. Wallach, H.~Larochelle, K.~Grauman,
  N.~Cesa{-}Bianchi, and R.~Garnett, Eds., 2018, pp. 5776--5787. [Online].
  Available:
  \url{https://proceedings.neurips.cc/paper/2018/hash/aeefb050911334869a7a5d9e4d0e1689-Abstract.html}
\BIBentrySTDinterwordspacing

\bibitem{huang2020multitask}
H.~Huang, Y.~Song, Y.~Wu, J.~Shi, X.~Xie, and H.~Jin, ``Multitask
  representation learning with multiview graph convolutional networks,''
  \emph{IEEE Transactions on Neural Networks and Learning Systems}, 2020.

\bibitem{buffelli2020meta}
D.~Buffelli and F.~Vandin, ``A meta-learning approach for graph representation
  learning in multi-task settings,'' \emph{arXiv preprint arXiv:2012.06755},
  2020.

\bibitem{lu2011link}
L.~L{\"u} and T.~Zhou, ``Link prediction in complex networks: A survey,''
  \emph{Physica A: statistical mechanics and its applications}, vol. 390,
  no.~6, pp. 1150--1170, 2011.

\bibitem{mccallum2000automating}
A.~K. McCallum, K.~Nigam, J.~Rennie, and K.~Seymore, ``Automating the
  construction of internet portals with machine learning,'' \emph{Information
  Retrieval}, vol.~3, no.~2, pp. 127--163, 2000.

\bibitem{tarkowski2016closeness}
M.~Tarkowski, P.~Szczepa{\'n}ski, T.~Rahwan, T.~Michalak, and M.~Wooldridge,
  ``Closeness centrality for networks with overlapping community structure,''
  in \emph{Proceedings of the AAAI Conference on Artificial Intelligence},
  vol.~30, no.~1, 2016.

\bibitem{nagaraja2010impact}
S.~Nagaraja, ``The impact of unlinkability on adversarial community detection:
  effects and countermeasures,'' in \emph{International Symposium on Privacy
  Enhancing Technologies Symposium}.\hskip 1em plus 0.5em minus 0.4em\relax
  Springer, 2010, pp. 253--272.

\bibitem{sen2008collective}
P.~Sen, G.~Namata, M.~Bilgic, L.~Getoor, B.~Galligher, and T.~Eliassi-Rad,
  ``Collective classification in network data,'' \emph{AI magazine}, vol.~29,
  no.~3, pp. 93--93, 2008.

\bibitem{borgwardt2005protein}
K.~M. Borgwardt, C.~S. Ong, S.~Sch{\"o}nauer, S.~Vishwanathan, A.~J. Smola, and
  H.-P. Kriegel, ``Protein function prediction via graph kernels,''
  \emph{Bioinformatics}, vol.~21, no. suppl\_1, pp. i47--i56, 2005.

\bibitem{dobson2003distinguishing}
P.~D. Dobson and A.~J. Doig, ``Distinguishing enzyme structures from
  non-enzymes without alignments,'' \emph{Journal of molecular biology}, vol.
  330, no.~4, pp. 771--783, 2003.

\bibitem{schomburg2004brenda}
I.~Schomburg, A.~Chang, C.~Ebeling, M.~Gremse, C.~Heldt, G.~Huhn, and
  D.~Schomburg, ``Brenda, the enzyme database: updates and major new
  developments,'' \emph{Nucleic acids research}, vol.~32, no. suppl\_1, pp.
  D431--D433, 2004.

\bibitem{wale2008comparison}
N.~Wale, I.~A. Watson, and G.~Karypis, ``Comparison of descriptor spaces for
  chemical compound retrieval and classification,'' \emph{Knowledge and
  Information Systems}, vol.~14, no.~3, pp. 347--375, 2008.

\bibitem{yanardag2015deep}
P.~Yanardag and S.~Vishwanathan, ``Deep graph kernels,'' in \emph{Proceedings
  of the 21th ACM SIGKDD international conference on knowledge discovery and
  data mining}, 2015, pp. 1365--1374.

\bibitem{newman2006finding}
M.~E. Newman, ``Finding community structure in networks using the eigenvectors
  of matrices,'' \emph{Physical review E}, vol.~74, no.~3, p. 036104, 2006.

\bibitem{von2002comparative}
C.~Von~Mering, R.~Krause, B.~Snel, M.~Cornell, S.~G. Oliver, S.~Fields, and
  P.~Bork, ``Comparative assessment of large-scale data sets of
  protein--protein interactions,'' \emph{Nature}, vol. 417, no. 6887, pp.
  399--403, 2002.

\bibitem{mcauley2012learning}
J.~J. McAuley and J.~Leskovec, ``Learning to discover social circles in ego
  networks.'' in \emph{Advances in neural information processing systems}, vol.
  2012.\hskip 1em plus 0.5em minus 0.4em\relax Citeseer, 2012, pp. 548--56.

\bibitem{waniek2018hiding}
M.~Waniek, T.~P. Michalak, M.~J. Wooldridge, and T.~Rahwan, ``Hiding
  individuals and communities in a social network,'' \emph{Nature Human
  Behaviour}, vol.~2, no.~2, pp. 139--147, 2018.

\bibitem{goodfellow2014generative}
I.~Goodfellow, J.~Pouget-Abadie, M.~Mirza, B.~Xu, D.~Warde-Farley, S.~Ozair,
  A.~Courville, and Y.~Bengio, ``Generative adversarial nets,'' \emph{Advances
  in neural information processing systems}, vol.~27, 2014.

\bibitem{wang2020gcn}
X.~Wang, M.~Zhu, D.~Bo, P.~Cui, C.~Shi, and J.~Pei, ``Am-gcn: Adaptive
  multi-channel graph convolutional networks,'' in \emph{Proceedings of the
  26th ACM SIGKDD International Conference on Knowledge Discovery \& Data
  Mining}, 2020, pp. 1243--1253.

\end{thebibliography}

\end{document}